\documentclass[sigconf]{acmart}

\AtBeginDocument{%
  \providecommand\BibTeX{{%
    \normalfont B\kern-0.5em{\scshape i\kern-0.25em b}\kern-0.8em\TeX}}}

\usepackage{graphicx}
\usepackage{tabularx}
\usepackage{enumitem}
\usepackage{multirow}
\usepackage{array}
\usepackage{pifont}
\usepackage{pgfplots}
\usepackage{subcaption}
\usepackage{amsthm}
\usepackage{nicefrac}
\usepackage{balance}
\usepackage{amsmath}
\usepackage{algorithmic}
\usepackage{caption}
\usepackage{amsfonts}
\usepackage{array}
\usepackage{hyperref}
\usepackage[linesnumbered,ruled,vlined]{algorithm2e}

\newcommand{\model}{\textsf{GraphFC}}

\newcolumntype{+}{!{\vrule width 1.5pt}}

\begin{document}

\title{GraphFC: Customs Fraud Detection with Label Scarcity}


\author{Karandeep~Singh}
\authornote{Both authors contributed equally to this research.}
\affiliation{%
  \institution{Institute for Basic Science}
  \city{Daejeon}
  \country{Korea}
}
\email{ksingh@ibs.re.kr}

\author{Yu-Che~Tsai}
\authornotemark[1]
\affiliation{%
  \institution{National Taiwan University}
  \city{Taipei}
  \country{Taiwan}}
\email{roytsai27@gmail.com}

\author{Cheng-Te~Li}
\affiliation{%
  \institution{National Cheng Kung University}
  \city{Tainan}
  \country{Taiwan}}
\email{chengte@ncku.edu.tw}

\author{Meeyoug~Cha}
\affiliation{
  \institution{Data Science Group, IBS \& School of Computing, KAIST}
  \city{Daejeon}
  \country{Korea}
}
\email{meeyoungcha@kaist.ac.kr}

\author{Shou-De~Lin}
\affiliation{%
 \institution{National Taiwan University}
 \city{Taipei}
 \country{Taiwan}}
 \email{sdlin@csie.ntu.edu.tw }

\renewcommand{\shortauthors}{Singh and Tsai, et al.}


\begin{abstract}
Custom officials across the world encounter huge volumes of transactions. With increased connectivity and globalization, the customs transactions continue to grow every year. Associated with customs transactions is the customs fraud - the intentional manipulation of goods declarations to avoid the taxes and duties. With limited manpower, the custom offices can only undertake manual inspection of a limited number of declarations. This necessitates the need for automating the customs fraud detection by machine learning (ML) techniques. Due the limited manual inspection for labeling the new-incoming declarations, the ML approach should have robust performance subject to the scarcity of labeled data. However, current approaches for customs fraud detection are not well suited and designed for this real-world setting. In this work, we propose \model{} (\textit{\textbf{Graph}} neural networks for \textit{\textbf{C}}ustoms \textit{\textbf{F}}raud), a model-agnostic, domain-specific, semi-supervised graph neural network based customs fraud detection algorithm that has strong semi-supervised and inductive capabilities. With upto 252\% relative increase in recall over the present state-of-the-art, extensive experimentation on real customs data from customs administrations of three different countries demonstrate that \model{} consistently outperforms various baselines and the present state-of-art by a large margin.

\end{abstract}



\keywords{Graph Neural Network, Frauds detection,  Multi-task learning
}

\maketitle

\section{Introduction}

\begin{figure}[ht!]
  \centering
  \includegraphics[width=1.0\linewidth]{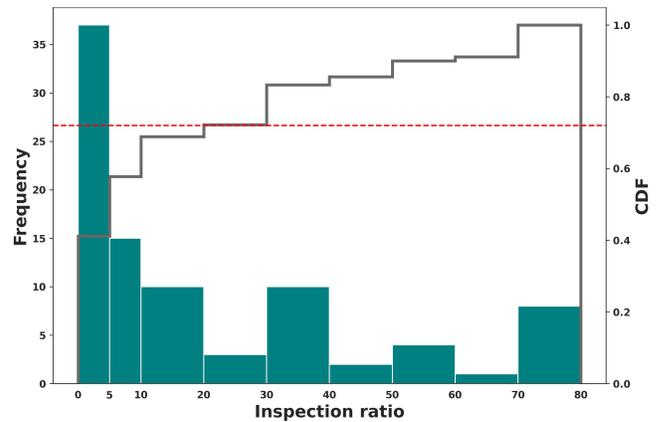}
  \caption{Distribution of physical inspection ratio among 90 countries across the world according to~\cite{worldbank}. Around 72\% countries have inspection rate less than 30\%.}
  \label{fig:inspect_rate}
  \vspace{-2mm}
\end{figure}

Customs is an authority in a country responsible for collecting tariffs and controlling the flow of goods into and out of country. According to the World Trade Organization, the world merchandise trade volume for exports and imports in 2019 alone was about 38 trillion dollars\footnote{Source: World Integrated Trade Solution,\\ \url{https://wits.worldbank.org/CountryProfile/en/WLD}}. With globalization and increased connectivity, the trade volumes continue to grow further. 

Customs cover a wide range of trade facilitation issues aimed at greater transparency, effectiveness, and efficiency of government services and regulations with regard to the clearance of import, export and transit transactions across international borders. 
In addition to these aspects, customs also include facilitation measures relating to cross-border trade and supply-chain security.
International trade and commerce via customs encounter malicious transaction declarations that involve intentional manipulation of trade invoices to avoid ad valorem taxes and duties~\cite{web1,web2,web3}. Administrators inspect trading goods and invoices to secure revenue generation from trades and develop automated systems to detect fraudulent transactions. This is an important undertaking as with customs transactions running into millions, even a slight increase in the detection accuracy will result in collection of hundreds of thousands of dollars of additional revenue. However, limited manpower and the colossal scale of trade volumes make the manual inspection of all transactions practically infeasible. In fact, as per the World Bank reports, administrations end up inspecting only a small portion, typically 5\% or lower, of the total transactions~\cite{geourjon2013inspecting,worldbank}.
To select the most suspicious transactions to be inspected, customs administrators often leverage rule-based algorithms or expert-systems with the domain know-how and empirical knowledge to determine the illicitness of transactions~\cite{vanhoeyveld2020customs}. Recently, the World Customs Organization (WCO)\footnote{\url{http://www.wcoomd.org/}} and its partner countries have made a great deal of progress in developing machine learning models, which can help officials identify and investigate any suspicious transactions that yield maximum revenue. 
For instance, India and China Customs Services applied decision tree and neural network based algorithms that outperform the traditional rule-based system~\cite{shao2002applying,kumar2006development}. Moreover, the WCO had released DATE~\cite{kimtsai2020date}, a deep learning-based approach that jointly considers the information of importers and their trade goods and achieves state-of-the-art performance in detecting fraudulent transactions.

\begin{figure}[t!]
\flushleft
\centering
\includegraphics[width=0.8\linewidth]{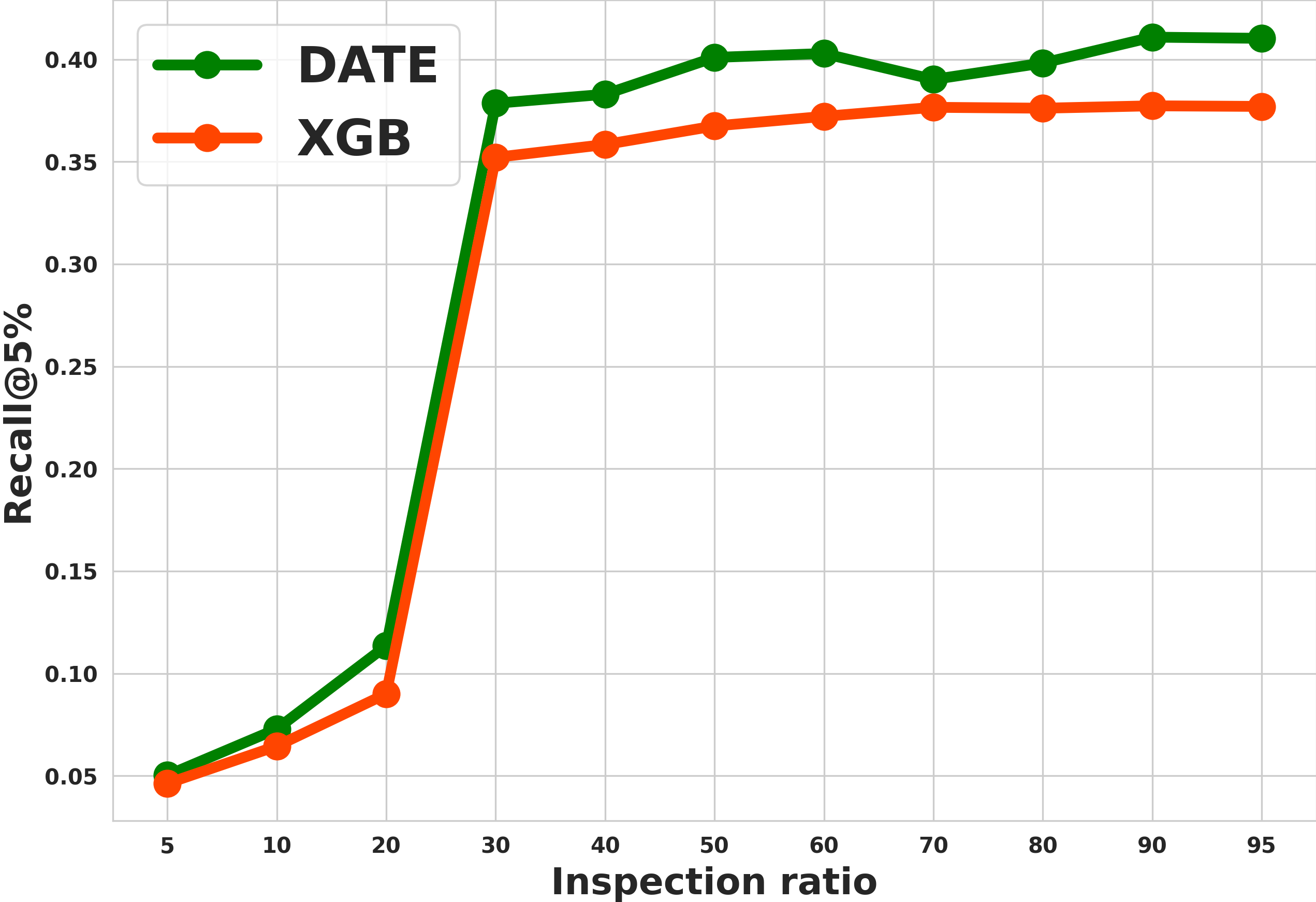}
\caption{Performance w.r.t. manual inspection ratio (\%).}
\label{fig:performance_wrtInspection}

\end{figure}

Although there are various approaches proposed for customs fraud detection, none of the existing work has addressed a critical issue: low inspection ratios and thereof lack of labeled ground-truth data for training the ML algorithms. 
 
Figure~\ref{fig:inspect_rate} presents the distribution of physical inspection rate of 90 different countries across the world. Intuitively, we notice that around 40\% of countries facilitate physical inspection less than 5\% and up to 60\% of countries has inspection rate less than 10\%. Therefore, with low inspection ratio comes with the label scarcity. 
Scarcity of labeled data could severely hamper the performance of existing ML models for customs fraud detection models. We notice that most of the previous works and DATE~\cite{kimtsai2020date,shao2002applying,kumar2006development}, focus on a supervised learning paradigm that requires an adequately large amount of labeled data, which might become infeasible or less effective in fraud detection when dealing with limited labeled data. 

To understand the impact of performance of existing approaches under different inspection ratios, we randomly sample n\% of transactions from a fully labeled dataset ($\sim$2M labeled data) collected from a partner country of WCO (country B) and train supervised model learning models (e.g.,DATE and XGBoost). The results are presented in Figure~\ref{fig:performance_wrtInspection}.
We notice that supervised models achieve reasonably stable performance when the inspection rate is greater than 30\%. However, in the case of inspection ratio lower than 30\%, large number of data points become unlabeled and the performance drops significantly. 

While the performance of a supervised ML model is limited by the scarcity of labeled data, the potential of utilizing heaps of unlabeled data in the customs domain has not been explored yet. Thus, in this work, we aim to mitigate the problem of label scarcity for ML driven customs fraud detection approaches by adopting a \emph{semi-supervised} modeling approach.

To mitigate the difficulties arising from low inspection ratios, we propose a graph neural network (GNN) based customs fraud detection model and tackle this problem from the graph perspective.
The setup can be viewed as a graph where the individual transactions act as nodes and form the network, and a pre-defined connectivity measure connects nodes, forming links in the network. 
For instance, if transaction $txn_1$ was made by importer $imp_1$ with the item coded $hs_1$, and importer id and item codes are defined to be the connectivity measures, links can be formed between transaction nodes as shown in Figure~\ref{fig:graph_illus}).

GNNs have been proven to have strong representational abilities in the semi-supervised setting~\cite{gcn, graphsage, pinsage, gat}.  Moreover, self- and unsupervised pretraining approaches in GNNs enable learning of hidden relations and dynamics of interacting entities~\cite{g2g, gnninfomax, kipf2016variational}. 
While learning the transactions representations, supervised approaches like XGBoost and DATE are not only incapable of such an unsupervised pretraining, but also of accommodating the information from neighboring transactions. In customs setting, the representations of transactions learnt by GNNs are enriched by amalgamating information from the linked transactions as well as the structure of the network.

Motivated by this, we propose \textit{\textbf{Graph}} neural networks for \textit{\textbf{C}}ustoms \textit{\textbf{F}}raud (\model{}), a GNN-based customs fraud detection model operating on the structured customs data. The strength of \model{} lies in its two-stage design. In the first stage, we employ self-supervised learning technique that jointly learns the representation among labeled and unlabeled data. The self-supervision process allows \model{} understands the distribution of transactions and its related transactions throughout the message passing operation of GNN and the higher order connectivity in the transaction graph. Afterwards, in the second stage, we fine-tuned \model{} with labeled data under a dual-task learning scheme which predicts the illicitness and the expected revenue simultaneously to make sure the most suspicious and valuable transaction could be detected and retrieved. The key finding of leveraging GNNs for fraud detection is it not only learns the knowledge from unlabeled data through self-supervised learning, but also capable of referring other associated transaction when making predictions. For instance, tradition models predict the illicitness of $txn_1$ solely based on the features derived from $txn_1$, while \model{} takes the features of $txn_1$ and its relevant transaction $txn_2, txn_3$ into account. We found that such design could substantially ease the performance drop with label scarcity and offer major improvements over strong baselines like XGBoost and the present state-of-the-art DATE model~\cite{kimtsai2020date}. The model code is made publicly available. We summarize our contributions as follows:
\begin{itemize}[leftmargin=*]
    \item We develop \textit{\textbf{Graph}} neural networks for \textit{\textbf{C}}ustoms \textit{\textbf{F}}raud (\model{}) and showcase its strength in real-world customs fraud detection.
    \item \model{} adopts self-supervised and semi-supervised learning techniques to extract rich information from unlabeled data, which substantially alleviate the performance degradation due to label scarcity.
     \item \model{} optimizes on the dual of maximizing the detection of fraudulent transactions and maximizing the collection of associated revenue. Hence, it prioritizes fraudulent items that are likely to yield maximum penalty surcharge.
    \item Extensive experimentation on multi-year customs data from three countries shows that~\model{} outperforms the previous state-of-the-art and various other baselines.
\end{itemize}

\section{Related Work}

Most of the customs administrations are using legacy rule-based systems~\cite{Kultur}. Perhaps the biggest advantage of rule-based systems is their simplicity and interpretability. But rule systems are hard to maintain and are heavily dependent on expert knowledge~\cite{Pozzolo, KRIVKO20106070}. In general, advances in data science have led to the development of various fraud-detection models. A popular choice is tree-based models~\cite{ABDALLAH201690,WEST201647,adewumi2017survey,AHMED2016278}.

Due to the non-availability of customs data in the public domain, the published literature on customs fraud is understandably scarce.
However, there are some known efforts, like Belgium, where customs have tested an ensemble method of a support vector machine-based learner in a confidence-rated boosting algorithm~\cite{vanhoeyveld2019customs}; Columbia has demonstrated the use of unsupervised spectral clustering in detecting tax fraud with limited labels~\cite{deRoux2018taxkdd}; Indonesia customs have proposed an ensemble of tree-based approaches, support vector machine and neural network~\cite{Canrakerta_2020}. Similarly, Netherlands' customs fraud detection model is built based on the Bayesian network and neural networks~\cite{TRIEPELS2018193}. Another research employed a deep learning model to segregate high-risk and low-risk consignment on randomly selected 200,000 data from Nepal Customs of the year 2017~\cite{Regmi2018}. Other countries have developed their customs fraud detection systems such as Brazil~\cite{jambeiro08jmlr}.

GNNs are being used in fraud detection. For instance~\cite{gnnminigraph} addressed the homogeneity and heterogeneity of issue of networks for fraudulent invitation detection, anomaly detection on dynamic graphs with sudden bouts of increased and decreased activities is proposed in~\cite{gnnanomaly},~\cite{gnn_semifatfraud2019} proposed a semi-supervised approach and an attention-based approach for fraud detection in Alipay. In~\cite{rao2020suspicious}, authors designed a dynamic heterogeneous graph from the users' registrations to prevent the suspicious massive account registrations,~\cite{gnncamouflaged2020} designs a GNN based fraud detection model to detect the camouflaged fraudsters,~\cite{fdgars2019} designs a graph convolutional network for fraudster detection in online app review system,~\cite{rao2020xfraud} undertakes the task of fraud detection in credit cards by learning temporal and location-based graph features.

Graph structure can also be defined as homogeneous, where nodes have the same set of features, such as the original GCN (Graph convolution network)~\cite{gcn}, where all the nodes (club members) have identical set features or are heterogeneous with nodes possessing a different set of features~\cite{rao2020xfraud, schlichtkrull2018modeling}. While these approaches made advances in the GNN methodologies and applications, their application in customs domain is completely unexplored.

\section{Problem Setting}

\subsection{Customs Selection}
\label{lab:probsettin}
There are variety of customs fraud, but undervaluation is the most common type. Importers or exporters declare the value of their trade goods lower than the actual value, mainly to avoid ad valorem customs duties and taxes~\cite{rodrigue2016geography}. For this work, we limit the definition of an illicit customs transaction to that of undervaluation. 
Given the volume of transactions and limited resources to inspect them, the customs administration prioritizes items that entail a maximum penalty. In this light, we formulate the problem of customs selection as follows: 

\noindent \textbf{Problem: }\emph{Given a transaction $o_i$ with its importer ID $imp_i$ and HS-code
$c_i$ of the goods, the goal is to predict both the fraud score $\hat{y}_i^{cls}$ and the raised revenue $\hat{y}_i^{rev}$ obtainable by inspecting transaction $o_i$.}

By selecting two predicted values, $\hat{y}_i^{cls}$ and $\hat{y}_i^{rev}$ in descending order from all transactions $O = \{o_1, \ldots, o_n\}$
, customs administration could identify the most suspicious transactions 
$O_F\subset O$ to be inspected. 
Meanwhile, since we focus on mitigating the low inspection ratio problem in this work, we assume we have the labeled set $L = \{(x_i,y_i^{cls},y_i^{rev}), \forall i=1,\ldots,n\}$ and unlabeled set $U=\{(x_i), \forall i = n+1,\ldots,m\}$, where $|L| \ll |U|$. 

Note that we also exhibit the model performance in the supervised setting and leave the results in Appendix, although a very high inspection rate( $>$ 90\%) is a rarity for customs administrations.

\subsection{Transactions Graph from Tabular Data}
\label{prob:fraudingraph}
\begin{figure}
  \centering
  \includegraphics[width=1\linewidth]{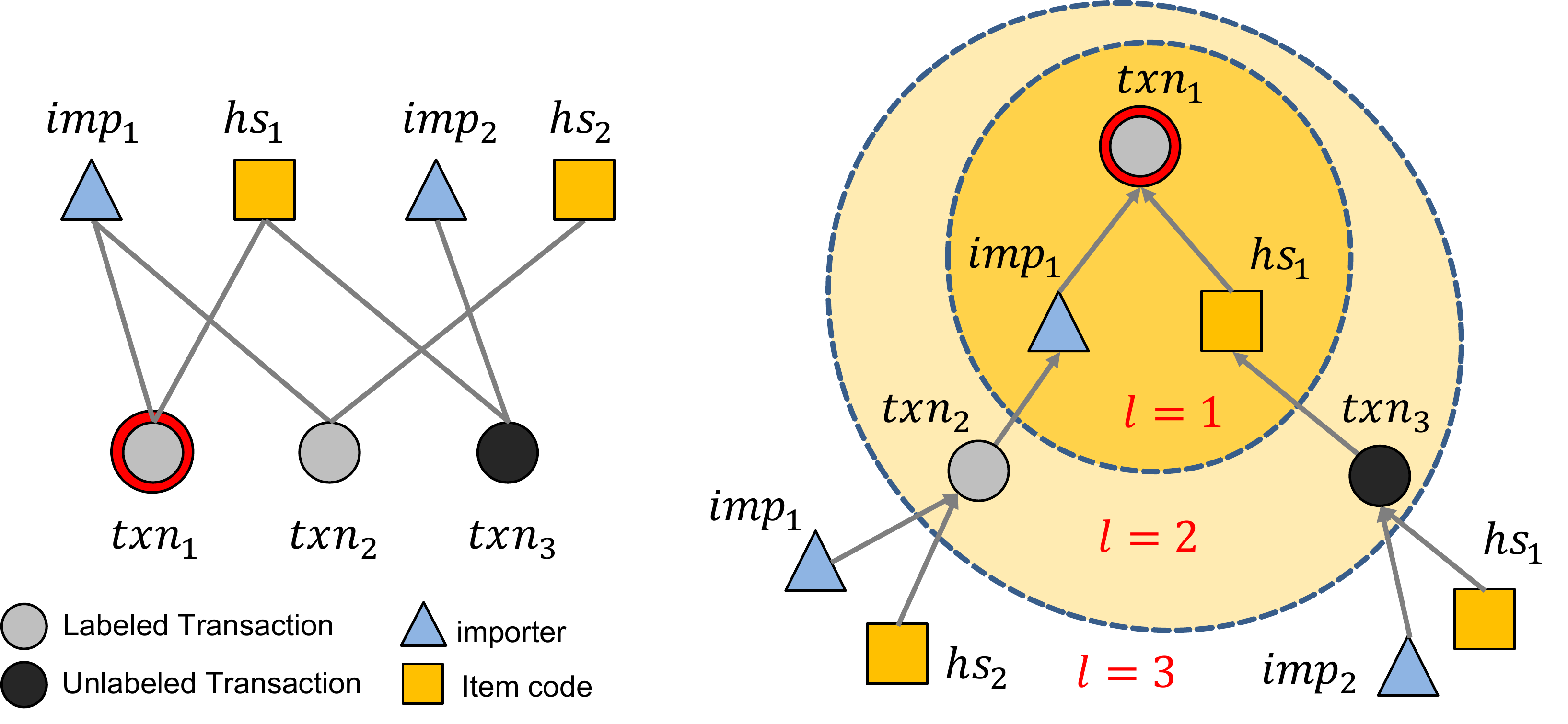}
  \caption{Transaction graph construction in \model{}}
  \label{fig:graph_illus}
\end{figure}

\begin{figure*}[!t]
  \centering
  \includegraphics[width=1\linewidth]{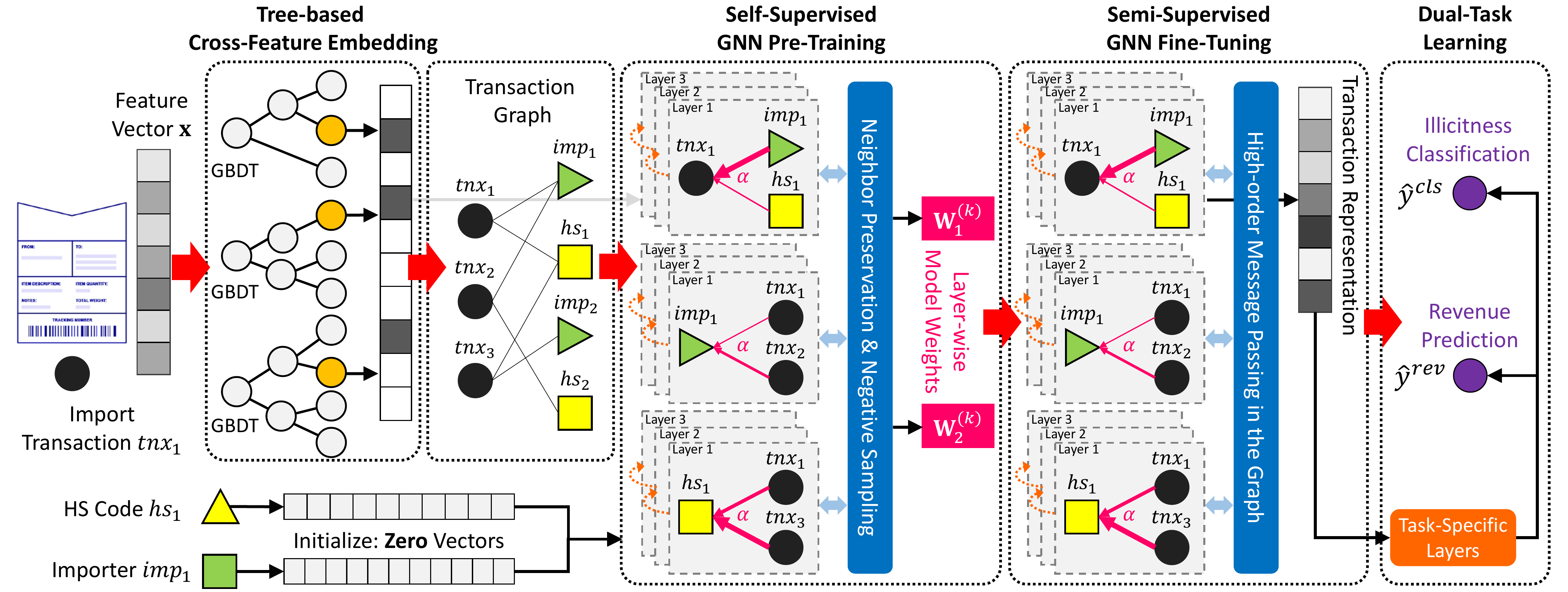}
  \caption{Model architecture of \model{}. Cross features extracted from GBDT step act as node features in the transaction graph. In the pre-training stage, \model{} learns the model weights and refine the transaction representations. Afterwards, the model is fine-tuned with labeled data with dual-task learning framework to predict the illicitness and the additional revenue.}
  \label{fig:architecture}
\end{figure*}

One of the key insights of this work is to represent the customs transactions and their interactions as a graph and use GNNs to learn transaction-level embeddings by aggregating the neighborhood information. Traditional fraud detection methods~\cite{kimtsai2020date,vanhoeyveld2019customs,jambeiro08jmlr}
judge the illicitness of a transaction solely depending on its features $x_i$ while the information of similar transactions is not considered.

We set up links (edges) between transactions (nodes) to capture any implicit patterns they might exhibit. These links could either be based on common categorical features, binned numerical features, or any other pre-defined rule. 
However, with even a relatively smaller customs office running into millions of transactions, we immediately run into a critical issue to exploding space complexity. A homogeneous graph formed by above rule suffers from a complexity of $O(n^2)$ due to the pairwise relational nature, where $n$ denotes the number of transactions. 
To handle this issue, we introduce a set of \emph{virtual nodes} $\mathcal{V_C}$ that are initially represented by the uniquely identifiable values/features (such as the importer id). Each node in $\mathcal{V_C}$ will only connect to transactions that share \textit{that} feature value with it, and thus bringing down the space complexity from $O(n^2)$ to $O(n)$. 

Based on previous works like DATE~\cite{kimtsai2020date}, we choose importer ID and HS-code to be the categorical features defining the links between transactions. Thus, in this work, $\mathcal{V_C}$ is composed of the importer ID and HS-code with the uniquely identifiable values as their node IDs, as illustrated in the left half of Fig.~\ref{fig:graph_illus}. In line with this, when a prediction is made for a transaction $\bf{txn_1}$, the features of other transactions that share the same importer ID($\bf{txn_2}$) and HS-code($\bf{txn_2}$) would be propagated to $\bf{txn_1}$ to form a new representation, as demonstrated in the right side of Fig.~\ref{fig:graph_illus}.
The detailed propagation rule will be described in Sec.~\ref{sec:method}. The transaction nodes $\mathcal{V_T}$ consist of both labeled and unlabeled transactions, which benefits the aggregation of additional information under a self- and semi-supervised setting.

The transaction graph could be represented as $\mathcal{G=(V,E)}$ where $\mathcal{V = V_T \cup V_C}$ is the node-set that denotes the combined individual transactions and the virtual nodes, and $\mathcal{E}$ denotes the connections between them based on the shared feature values. Let $\mathbf{D}\in \mathbb{R}^{h\times m}$ be the feature matrix that consists of $\mathbf{h=|\mathcal{V}|}$ 
nodes with their associated $\mathbf{m}$ features. It is worth noting that in the testing phase, only the historical transactions are considered for the inferring the transaction embedding. Also, the cross features from GBDT step become the features of nodes in $\mathcal{V_T}$, and the features of nodes in $\mathcal{V_C}$ are initialized with zero vectors $\vec{0} \in \mathbb{R}^{m}$.

\section{Methodology}
\label{sec:method}

The proposed model \model{}
synergizes the strengths the GBDTs and GNNs.
Model architecture for \model{} is illustrated in Fig.~\ref{fig:architecture}, highlighting its multi-stage design. 
\footnote{Model code: \url{https://github.com/k-s-b/gnn_wco}. Due to its proprietary nature, the raw data cannot be shared. Synthetic customs data for experimentation is provided.}

\subsection{Cross Features from GBDT}
Tree-based methodologies are a popular choice for building machine learning models on tabular data. Tabular data is known to have hyperplane-like decision boundaries, and tree-based models allow for hyperplane-based recursive partitioning of data space, such that points in the partitioned space share the same class.  
In this work, we follow the feature extraction method using GBDTs as~\cite{kimtsai2020date}.
Gradient Boosting Decision Trees (GBDTs) methodology can be understood as an ensemble technique where multiple weak learner models are combined to provide a powerful overall learner. The decision path in the fitted trees could be seen as a new set of features, with each path representing a new cross feature made from the original features. 
Assume $C$ and $E_T$ respectively represent internal nodes and their connecting paths in a decision tree $D_T$.
$C$ further can be viewed as consisting of three sets of nodes, the root node $\{c_R\}$, internal nodes $C_I$, and the leaf nodes $C_E$, where $C=\{c_R\}\cup C_I\cup C_E$. Nodes in $c_I\in C_I$ splits features into some decision space. After passing through internal nodes, an input feature vector $\mathbf{x} \in \mathbf{X}$ gets assigned to a leaf node $l_e$. Then, a \textit{decision rule} is the path connecting the root node to the leaf node.
For instance, ``\textsf{gross.weight \textgreater 100kg} $\land$ \textsf{quantity \textless~5}'' is a two-node decision path or a \textit{cross-feature} comprising of features \textsf{gross.weight} and  \textsf{quantity}. 

Inspired by some recent studies such as~\cite{wang2018www,he2014adkdd}, we utilize state-of-the-art XGBoost model~\cite{tianqi2016} and extract cross features from the fitted model.

Let us assume $W$ represents the total number of leaves in the ensemble. An input vector $\mathbf{x}$ ends up in one of the leaf nodes according to the decision rules of each fitted decision tree. Each activated leaf node in the $t$-th decision tree can be represented as a one-hot encoding vector 
$\mathcal{F}_t$. We concatenate these together and produce a multi-hot vector $\mathbf{p} \in \mathbb{R}^{W}$, where $1$ indicates the activated leaves and $0$, the non-activated ones. After the extraction of cross features, we apply transformation for all data points to obtain a new feature matrix $\mathbf{M} \in \mathbb{R}^{n\times W}$.

\subsection{Transaction Interaction Learning with Message Passing}

The strengths of learning latent representations of transactions via graph neural networks are mainly threefold. First, given a transaction node\footnote{In context of embeddings, we use the terms ``node(s)'' and ``transaction(s)'' invariably hereafter.} $v_t \in \mathcal{V_T}$ and it's representation vector $m^k_t$ at $k^{th}$ layer, graph neural network updates $m^k_t$ with the feature vector of adjacent nodes (i.e., $m_j^k~\forall j \in N(v_t)$). Hence the nearby transactions in $\mathcal{G}$ could be included to form the representation for $v_t$. Second, the aggregation process takes the neighboring \textit{labeled} and \textit{unlabeled} transactions into account, a critical step in semi-supervised settings. Lastly, existing work~\cite{kimtsai2020date} fails to extract informative patterns if either a new importer or an unseen item made the transaction. 
On the contrary, the inductive GNN generalizes to unseen nodes by exploring the node neighborhood. We elaborate the detailed framework in the following paragraphs.

In this work, \model{} adopts GNN architecture inspired by GAT~\cite{velivckovic2017graph},
that has shown strong representation learning abilities among different graph learning tasks. 
Thanks to the flexible nature of our proposed approach, the backbone GNN architecture could be replaced with \textbf{any GNN model}. We demonstrate this by experimenting with GraphSAGE~\cite{hamilton2017inductive} and RGCN~\cite{schlichtkrull2018modeling} (refer Table~\ref{tab:ssl-results}).

We initialize the node embedding for $\mathcal{V_C}$ with zero vectors then update it via the propagation from neighboring transaction features.

\subsubsection{Local subgraph extraction via neighbor sampling}
\label{method:local_subgraph}
Training transductive GNNs such as GCN~\cite{gcn} usually requires loading the entire graph data and the intermediate states of all nodes into memory. The full-batch training algorithm for GNNs suffers significantly from the memory overflow problem, especially when dealing with large scale graphs containing millions of nodes. Also, the addition of new nodes would require retraining of the entire model. 
To mitigate these issues, the neighborhood sampling technique~\cite{hamilton2017inductive} is applied to extract a local subgraph from a target node for inductive mini-batch training. Specifically, it samples a tree rooted at each node by recursively expanding
the root node’s neighborhood by $K$ steps with a fixed sample size of $\{T_1, T_2,...,T_K\}$, where $T_i$ is the number of nodes sampled in $i$-th step. Note that we alternately sample $T_i$ nodes from either $\mathcal{V_C}$ or $\mathcal{V_T}$ at each step to form the subgraph. Afterwards, the desired message passing operation could be applied to the subgraph.

\subsubsection{Message passing with graph neural network }
\label{method:message_passing}
Intuitively, the interaction between nodes can be considered as an embedding propagation (i.e., \emph{message passing}) from adjacent nodes to the source node. By stacking multiple message passing operations, the node representations collect information from a local subgraph, as illustrated in Fig.~\ref{fig:graph_illus} and Sec.~\ref{prob:fraudingraph}.

The process of message construction derives the joint representation of a node by passing its features through a neural network. The message aggregation process updates all of the neighboring messages obtainable from the previous message construction stage. It unifies them into a single representation as the final node embedding.

\textbf{Message Construction.} 
Given set of node pair $(m,j)$, where one of $\{m,n\}$ falls in $\mathcal{V_T}$ and the other belongs to $\mathcal{V_C}$. We first define the message from node $n$ to node $m$ as:
\begin{equation}
    g_{m\leftarrow j} = f(s_m,s_j),
    \label{method:firstorder}
\end{equation}
where $g_{m\leftarrow j}$ is the message embedding(i,e, information to be propagated) and $s_k$ denotes the $k^{th}$ row in $\mathbf{M}$. $f(\cdot)$ is the message encoding function which takes the node embeddings of $\{m,j\}$ and output their joint representation, $f(\cdot)$ is implemented as follows:
\begin{equation}
    f(s_m,s_j) = \alpha_{mj} s_j,
\end{equation}
where $\alpha_{mj}$ is the attention score that represents the similarity between nodes which is given as:
\begin{equation}
\label{GNN:attention_score}
    \alpha_{mj} = \frac{\exp\left(\sigma \left(\mathbf{r}^\top [\mathbf{W}_1 s_j \parallel \mathbf{W}_1 s_m]\right)\right)}{\sum_{n \in N(m)} \exp\left(\sigma\left(\mathbf{r}^\top [\mathbf{W}_1 s_j \parallel \mathbf{W}_1 s_j]\right)\right)},
\end{equation}
where $\mathbf{r}$ is the learnable vector that projects the embedding into a scalar, $\mathbf{\parallel}$ denotes the concatenation operation, $\sigma$ represents non-linear activation function(we use LeakyReLU~\cite{maas2013rectifier} in this work), and $\mathbf{W}_1$ is a learnable weight matrix.

It is worth pointing out here that different GNN methods have various strategies to discriminate edges when generating node embeddings.
In this work, we use attention-based aggregation to assign different importance weights to each edge so that the model can better distinguish transactions with different importers and with different HS-codes.

For example, RGCN~\cite{schlichtkrull2018modeling} exploits discriminate edges based on their relations, and SAGE~\cite{hamilton2017inductive} does not discriminate edges but treat all links equal.
 
Thus, any newer aggregation methodologies can be employed in future for enhancing the performance even further. 

\textbf{Message Aggregation.} In this stage, messages obtainable from node $m$ and all of it's neighboring nodes $\{j \in N(m) \}$ are aggregated into an unified representation. Aggregation function is defined as:
\begin{equation}
\label{eq:aggregation}
    s_m^{(1)} = ReLU\left(g_{m\leftarrow m} + \sum_{ j \in N(m)} \mathbf{W}_2~ g_{m\leftarrow j} \right),
\end{equation}
where $g_{m\leftarrow m} = W_2~s_m$ and 
$s_m^{(1)}$ denotes the new representation of node $m$ after the first propagation process and $W_2$ is a learnable weight matrix.\\
Given the first-order propagation rule (i.e., Eq.~\ref{method:message_passing}), we could further capture the higher-order transaction signals by stacking $k$ embedding propagation layer. 
Formally, the node representation at $k$-th layer could be derived as:
\begin{equation}
\begin{split}
\label{eq:highorder_aggregation}
     s_m^{(k)} &= ReLU\left(g_{m\leftarrow m}^{(k)} + \sum_{ j \in N(m)} \mathbf{W}_2^{(k)}~ g_{m\leftarrow j}^{(k)} \right),\\
     &= ReLU\left(\sum_{ j \in N(m)\cup\{m\}} \mathbf{W}_2^{(k)} \alpha_{mj}^{(k)}s_j^{(k-1)} \right)
\end{split}
\end{equation}

where $W_2^{(k)}$ is a trainable matrix and $\alpha_{mj}^{(k)}$ is defined similar as Eq.~\ref{GNN:attention_score} which $W_1$ is replaced by $W_1^{(k)}$ denotes the learnable weight matrix at the $k$-th layer. In this work, we use the embeddings at last layer (i.e., $s_m^{(k)}$) as the final representation for each transaction.

\subsubsection{Order of Message Passing}
In the message passing operation, one of the key components of \model{} is to make use of the virtual node $\mathcal{V_C}$ to aggregate features from its neighboring transactions, with $\mathcal{V_C}$ being connected to both labeled and unlabeled transactions.
As the node embeddings of virtual node $\mathcal{V_C}$ is initialized with zero vectors, it is critical to decide the order of message passing to avoid collecting information from an empty vector. We address this issue by updating embeddings of $\mathcal{V_C}$ 
and $\mathcal{V_T}$ in the following manner:
In step 1, the embeddings for $\mathcal{V_C}$ are updated by collecting information from $\mathcal{V_T}$ via E.q~\ref{method:firstorder}. It is followed by step 2 that updates  embeddings of $\mathcal{V_T}$ as per new representations of $V_C$. Step 1 and step 2 performed alternately to obtain the final embeddings.

\subsection{Self-Supervised Pretraining}

Unsupervised pretraining approaches in GNNs
enable learning of hidden relational patterns and structural properties of the underlying network. The fact that such information can be learnt solely from observational data like node and edge features, without the need for ground-truth labels, lends an amazing flexibility to the representational learning of the GNNs~\cite{g2g, gnninfomax, kipf2016variational}. Inspired by this, we adopt a self-supervised GNN pretraining scheme that learns transactions embeddings by using the given features of both the labeled and unlabeled transactions data.. Thereafter, we utilize these embeddings as a prior for the semi-supervised learning stage of \model{}.
The objective of the self-supervised step is to learn the embeddings of a node such that its representation in the latent space is closer to its neighboring nodes. By forcing the embedding of nearby nodes closer, GNN learns to preserve the graph structure level information as discussed in~\cite{grover2016node2vec}. Besides, it improves the generalization ability by aligning the distribution of unlabeled data closer to labeled data and thus reduce the extrapolation phenomenon.

We utilize the following self-supervised pretraining objective function~\cite{hamilton2017inductive}:
\begin{equation}
\begin{split}
    \mathcal{L}_{pre}(s_u^{(k)}) &= \frac{-1}{|N(u)|} \sum_{v\in N(u)} \log \left( \phi \left( s_u^{(k)\top} s_v^{(k)} \right) \right) \\
     &+ \frac{1}{R} \sum_{v_n \in P_n} \log \left( \phi \left( s_u^{(k)\top} s_{v_n}^{(k)} \right) \right)
    \label{loss:pretrain}
\end{split}
\end{equation}
where $v$ is the immediate neighbor of node $u$, $P_n$ denotes the negative sampling distribution that gives a set of negative sampled nodes, and $R$ is the number of negative samples. By minimizing the objective function in Eq.~\ref{loss:pretrain}, \model{} is forced to lower the distance of representation of nearby nodes while pushing away the negative sampled nodes. Note that the pretraining operation is applied for all nodes in $\mathcal{V}$ which implies that both labeled and unlabeled information would be used to learn robust GNN parameters. Lastly, the trained parameters are saved and fine-tuned on the labeled dataset with Eq.~\ref{eq:loss}.

\subsection{Prediction and Optimization}
To determine the illicitness of a given transaction, we propose to map the graph-based representation into scalars and train the neural network parameters via the following objective function. Inspired by~\cite{kimtsai2020date}, the dual-task learning framework is used to achieve two goals: 1. provide the probability of a transaction being illicit 2. predict the additional revenue (i.e., taxes) after inspecting suspicious transactions. Hence, we use the transaction representation (i.e., $s_m^{(k)}$) for both tasks of binary illicit classification and maximizing revenue prediction. Given the transaction feature $s_m^{(k)}$, we introduce the task-specific layer: 
\begin{equation}
\label{eq:prediction}
\begin{aligned}
    \hat{y}^{cls}(s_m^{(k)}) &= \phi\left(\mathbf{r}_1^\top s_m^{(k)} + \mathbf{b}_1\right), \\
    \hat{y}^{rev}(s_m^{(k)}) &=  \mathbf{r}_2^\top s_m^{(k)} +\mathbf{b}_2,
\end{aligned}
\end{equation}
where $\textbf{r}_1, \textbf{r}_2 \in \mathbb{R}^d$ denotes the hidden vectors of task-specific layers that project $s_m^{(k)}$ into the prediction tasks of binary illicitness and raised revenue, respectively. $\phi$ is the sigmoid function. $\hat{y}^{cls}(s_m^{(k)})$ is the predicted probability of a transaction being illicit, and $\hat{y}^{rev}(s_m^{(k)})$ is the predicted raised revenue value of a transaction. The final objective function $\mathcal{L}_{\model{}}$ is given by:
\begin{equation}
\begin{aligned}
    \mathcal{L}_{\model{}} = \mathcal{L}_{cls} + \alpha \mathcal{L}_{rev} + 
    \lambda \| \Theta \|^2,
\end{aligned}
\label{eq:loss}
\end{equation}
where $\Theta$ denotes all learnable model parameters, $\mathcal{L}_{cls}$(refer Eq.~\ref{eq:l_cls}) is the cross-entropy loss for binary illicitness classification, $\mathcal{L}_{rev}$ (refer Eq.~\ref{eq:l_cls}) is the mean-square loss for raised revenue prediction. The hyperparameter $\alpha$ is used to balance the contributions.
 
Finally, we use mini-batch gradient descent to optimize the objective function $\mathcal{L}_{\model{}}$, along with the \emph{Ranger}~\cite{Ranger} optimizer.

\section{Evaluation}
\subsection{Experimental Setup}

\begin{table}[!t]
\caption{Brief data description of the three datasets.}
\vspace{-3mm}
\label{tab:data_description}
\small
\centering
\begin{tabular}{|c|c|c|c|c|}
\hline
                    & \textbf{A-Data}       & \textbf{B-Data}       & \textbf{C-Data} \\ \hline

Starting Date	& 2016-01-01	& 2014-01-01	& 2017-01-01		\\ \hline
Ending Date	& 2019-01-01	& 2017-01-01	& 2019-01-01		\\ \hline
\# Transactions	& 1895222	& 1428397	& 2385367		\\ \hline
\# importers	& 8699	& 165202	& 132893		\\ \hline
\# HS codes	& 5491	& 5953	& 13387		\\ \hline

Overall Illicit rate	&  1.21\%	& 4.12\%	& 8.16\%		\\ \hline

\end{tabular}
\end{table}
We employed multi-year transaction-level import data from three partner countries, released by WCO for research usage. Because the data contains only the import transactions and those reported with undervaluation, we are only working with limited data from a larger dataset. Due to data confidentiality policies, we abbreviate these countries as A, B, and C. Subject to data availability, the three datasets have a different number of records, details of which are summarized in Table~\ref{tab:data_description}. Further, as data belongs to different customs administrations, it varies slightly in the features. Nonetheless, the most relevant features are the same across the datasets. Among others, these include the HS-code, importer ids, the notional value of the transaction, date, taxes paid, the quantity of the goods. We also include a binary feature representing historic fraudulent behavior or total taxes paid per unit value.

\vspace*{0.05in}
\noindent \textbf{Data Split.}
To mimic the actual setting, we split the train, valid, and test data temporally. Since we have multi-year data, we treat the data from the most recent year as the test, and the older data is split as train and validation sets. 
Because of our proposed design of introducing the virtual nodes for HS-codes and importer ids as presented in Sec~\ref{prob:fraudingraph}, the number of edges in the resultant graph structures of each of the datasets would be twice the number of transactions in it.

\vspace*{0.05in}
\noindent \textbf{Evaluation Criteria.}
The customs administrations can manually inspect only a limited number of transactions; we imitate this setting by evaluating the model performance on top-n\% (we demonstrate results for 1\%, and 5\%) of the suspicious transactions suggested by the model. Thus, e.g., precision at 1\% is equivalent to classification precision when top 1\% of transactions suggested by \model{} are \textit{manually inspected} and the ground-truth for those transactions is obtained. Same as DATE~\cite{kimtsai2020date}, we use precision (Pre.), recall (Rec.) as evaluation metrics to inspect the model performance. 
As \model{} supports the dual-task learning, we also report revenue (Rev.) collected by correct identification of the illicit transactions. This metric is reported as the \textit{ratio} of the total revenue collected if all fraudulent transactions were identified. In a live setting, an inspection of top-n\% items suggested by the model equates to customs officers examining the transactions that would yield the maximum revenue (refer Eq.~\ref{eq:rev_c}). 

We run an elaborate set of experiments that aim to answer the following: 
\textbf{(a)} Subject to the limited inspection rate by the customs officers, how well can \model{} identify the fraudulent transactions? 
\textbf{(b)} Similarly, what additional revenue can \model{} generate by inspection of items suggested by it? 
\textbf{(c)} How unlabeled data helps in semi-supervised learning? 
\textbf{(d)} Do different components in \model{} contribute to making better predictions?, and finally, \textbf{(e)} Is the model robust to different backbone networks and inspection rates.

\subsection{Performance Comparison}
\label{sec:performance_compare}

\model{} performance is compared with 4 baseline methods:

\begin{itemize}[leftmargin=*]
    \item \textbf{{XGBoost}}~\cite{tianqi2016}: XGBoost is a tree-based model which is widely used for modeling tabular data.

    \item \textbf{{Tabnet}}~\cite{arik2019tabnet}: Tabnet is a self-supervised learning-based framework especially designed for tabular data. 
    
    \item \textbf{VIME}~\cite{yoon2020vime}: VIME adopts   self-and semi-supervised learning techniques for tabular data modeling. 
    \item \textbf{{DATE}}~\cite{kimtsai2020date}: This fraud detection method which utilizes tree-aware embeddings and attention network. 
    
\end{itemize}

\vspace*{0.05in}

\begin{table}[!t]

\caption{Performance under 5\% inspection rate. 
The 5\% ground truth labels are used (95\% of labels are masked). n denotes the top n\% transaction suggested by the model. 
}
\label{tab:ssl-results}
\scriptsize
\centering
\resizebox{0.47\textwidth}{!}{
\begin{tabular}{|c|c|c|c|c|c|c|}
\hline

\multicolumn{7}{|c|}{\textbf{A-Data}} \\ \hline
& \multicolumn{3}{c|}{\textbf{n = 1\%}} & \multicolumn{3}{c|}{\textbf{n=5\%}} \\ \hline
Model & Pre. & Rec. & Rev. &  Pre. & Rec. & Rev.    \\ \hline
XGB  & 0.026 & 0.021 & 0.040 & 0.017 & 0.070 & 0.096  \\ \hline
Tabnet  & 0.031 & 0.024 & 0.031 & 0.016 & 0.061 & 0.049  \\ \hline
VIME  & 0.021 & 0.018 & 0.030 & 0.007 & 0.030 & 0.038  \\ \hline
DATE  & 0.045 & 0.037 & 0.056  & 0.033 & 0.129 & 0.187  \\ \hline
$\model{}_{\text{SAGE}}$ & 0.040 & 0.032 & 0.039 & 0.026 & 0.104 & 0.105  \\ \hline
$\model{}_{\text{RGCN}}$ & \textbf{0.050} & \textbf{0.040} & \textbf{0.064} & \textbf{0.039} & \textbf{0.152} & \textbf{0.189}  \\ \hline
\model{} & 0.045 & 0.035 & 0.057 & 0.030 & 0.118 & 0.170  \\ \hline

\multicolumn{7}{|c|}{\textbf{B-Data}} \\ \hline
XGB  & 0.151 & 0.061 & 0.109 & 0.045 & 0.092 & 0.184  \\ \hline
Tabnet  & 0.149 & 0.060 & 0.089 & 0.046 & 0.085 & 0.171  \\ \hline
VIME  & 0.064 & 0.024 & 0.052 & 0.043 & 0.087 & 0.208  \\ \hline
DATE  & 0.152 & 0.061 & 0.105  & 0.057 & 0.115 & 0.210  \\ \hline
$\model{}_{\text{SAGE}}$  & 0.327 & 0.132 & 0.143  & \textbf{0.201} & 0.405 & 0.305  \\ \hline
$\model{}_{\text{RGCN}}$  & 0.259 & 0.104 & 0.127  & 0.179 & 0.362 & \textbf{0.334}  \\ \hline
\model{} & \textbf{0.401} & \textbf{0.162} & \textbf{0.171} & 0.200 & \textbf{0.405} & 0.304 \\ \hline

\multicolumn{7}{|c|}{\textbf{C-Data}} \\ \hline
XGB  & 0.671 & 0.070 & 0.120  & 0.445 & 0.234 & 0.359  \\ \hline
Tabnet  & 0.715 & 0.050 & 0.081 & 0.452 & 0.231 & 0.334  \\ \hline
VIME  & 0.725 & 0.076 & 0.116 & 0.471 & 0.249 & 0.362  \\ \hline
DATE  & 0.803 & 0.085 & 0.158 & 0.472 & 0.249 & 0.380  \\ \hline
$\model{}_{\text{SAGE}}$ & 0.788 & 0.083 & 0.135 & 0.535 & 0.282 & 0.442  \\ \hline
$\model{}_{\text{RGCN}}$  & 0.819 & 0.086 & 0.146 & 0.525 & 0.277 & 0.424 \\ \hline
\model{} & \textbf{0.869} & \textbf{0.092} & \textbf{0.170} & \textbf{0.535} & \textbf{0.282} & \textbf{0.445}  \\ \hline

\end{tabular}
}
\end{table}

\begin{figure*}[t!]
\flushleft
\includegraphics[width=1.0\linewidth]{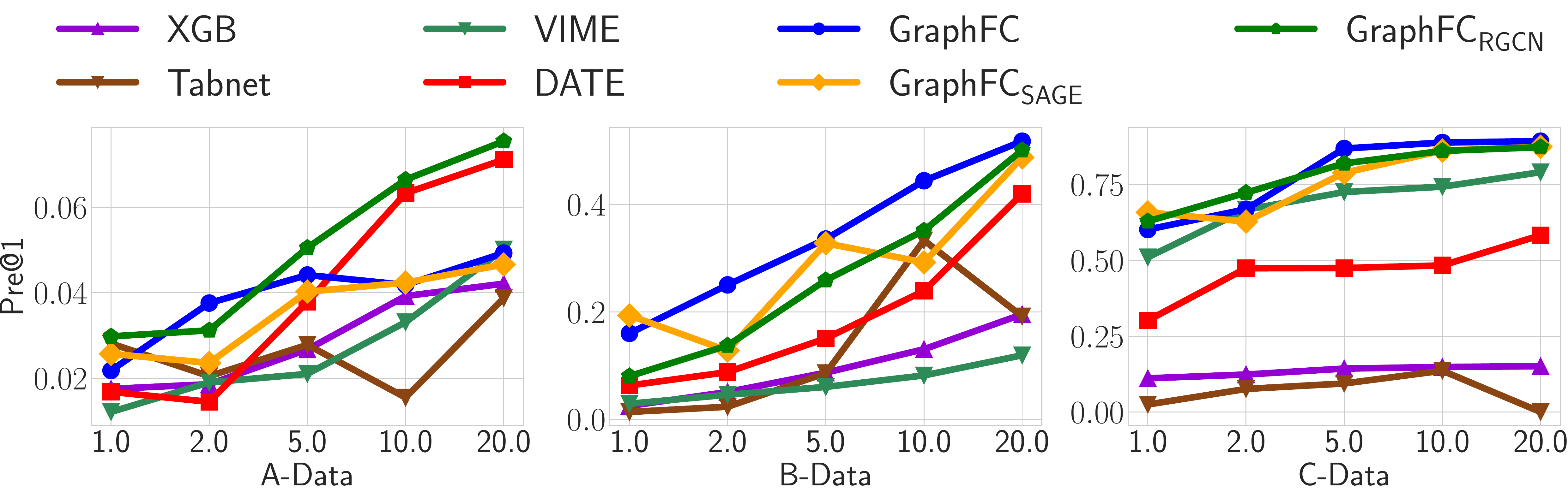}
\vspace{-4mm}
\caption{Performance comparison by varying inspection rate. The x-axis denotes the different inspection rate.}
\label{fig:varying_inspection}
\end{figure*}

\noindent \textbf{Performance under Low Inspection Ratio.}
To verify the effectiveness of the proposed model subject to label scarcity, 
we assume a 5\% inspection rate for each country and randomly sample 5\% of the transactions from the train set as labeled and mask labels for the remaining 95\% of the transactions. 
Results for this setting are presented in Table~\ref{tab:ssl-results}. Among the baseline methods, the current state-of-the-art customs fraud detection method, DATE, shows superiority due to its network design that considers the joint behavior between importer id and item codes. On the other hand, we notice that although VIME and Tabnet both utilize unlabeled data to perform self- and semi-supervised learning, the performance are still relatively lower than DATE and \model{}. The reason being design of both DATE and \model{} that explicitly consider the information of transactions made by different importers and trade goods.
\model{} shows a consistent and remarkable improvement over the baselines, which verifies the effectiveness of the proposed model. It is worth emphasizing that the significant improvement of \model{} over VIME and Tabnet further validates the superiority of the proposed model.

\noindent \textbf{Model Robustness.}
We demonstrate model robustness from two aspects: with different backbone GNNs, and by varying inspection rates. Specifically, we use the following backbone GNNs:

\begin{itemize}[leftmargin=*]
    
    \item \textbf{$\model{}_{\text{SAGE}}$}: A \model{} variant using GraphSAGE~\cite{hamilton2017inductive} aggregator in message passing. 
    \item \textbf{$\model{}_{\text{RGCN}}$}: A \model{} variant using RGCN~\cite{schlichtkrull2018modeling} aggregator designed for relational and heterogeneous graphs.
    
    \item \textbf{\model{}}: Proposed model with GAT~\cite{gat} based aggregator.
    
\end{itemize}

\noindent Inspection rates are varied in $\{1\%, 2\%, 5\%, 10\%, 20\%\}$, representing customs in different countries. Results with different backbone networks are presented in Table~\ref{tab:ssl-results}. Fig.~\ref{fig:varying_inspection} shows the model performance under different inspection rates. It can be observed that though the detection performance drops as the inspection rate decreases, \model{} and its variants still consistently outperform all the baselines. 
These results firmly establish that \model{} is robust against different degrees of label scarcity, and generalize well in different countries. Additionally, model robustness with different backbone GNNs is a proof that any desired aggregation strategy can be used in \model{} as per the features in the target data.

\subsection{Effectiveness of Unlabeled Data}
\begin{table}[t!]
\caption{Performance comparison on different training graphs under 5\% inspection rate. n denotes the top n\% transaction suggested by the model. }
\label{tab:ssl_variantgraph}
\small
\centering
\begin{tabular}{|c|c|c|c+c|c|c|c|}
\hline

\multicolumn{7}{|c|}{\textbf{B-Data}}                \\ \hline
      & \multicolumn{3}{c+}{\textbf{n = 1\%}} & \multicolumn{3}{c|}{\textbf{n=5\%}} \\ \hline
Model & Pre.     & Rec.    & Rev.    & Pre.    & Rec.    & Rev.   \\ \hline

Q($\mathcal{G}_L$) + F($\mathcal{G}_L$)  & 0.245   & 0.099  & 0.134  & 0.147  & 0.298  & 0.259 \\ \hline
Q($\mathcal{G}_{U}$) + F($\mathcal{G}_L$)  & 0.278   & 0.112  & 0.141  & 0.186  & 0.377  & 0.304 \\ \hline
Q($\mathcal{G}$) + F($\mathcal{G}_L$)  & 0.286   & 0.115  & 0.141  & 0.180  & 0.363  & 0.285 \\ \hline
Q($\mathcal{G}_L$) + F($\mathcal{G}$)  & 0.292   & 0.118  & 0.149  & 0.163  & 0.329  & 0.269 \\ \hline
Q($\mathcal{G}_{U}$) + F($\mathcal{G}$)  & 0.402   & 0.162  & 0.155  & 0.173  & 0.349  & 0.286 \\ \hline
Q($\mathcal{G}$) + F($\mathcal{G}$)  & \textbf{0.402} & \textbf{0.162}  & \textbf{0.172}  & \textbf{0.200}  & \textbf{0.406}  & \textbf{0.305} \\ \hline

\end{tabular}
\end{table}
\model{} improves its representation ability by utilizing the unlabeled data in the pretraining and fine-tuning stage. The key to making use of unlabeled data lies in the construction of transaction graph as presented in Fig.~\ref{fig:graph_illus}. The transaction graph comprises of $V_T$ and $V_C$, where $V_T$ includes both labeled and unlabeled transactions. To verify the
effectiveness of unlabeled data in $V_T$, two variants of $\mathcal{G}$ are made with the following rule: \textbf{(a)} $\mathcal{G}_L$: 
keeps only the labeled transaction and its corresponding edges with $V_C$. \textbf{(b)} $\mathcal{G}_{U}$: keeps only the unlabeled transaction and its corresponding edges with $V_C$. We then compare the results using different graphs in the pretraining(Q) and fine-tuning (F) stage
With 6 combinations and list the performance with a semi-supervised setting of B-Data in Table~\ref{tab:ssl_variantgraph}. Each row represents the graphs used in pretraining and fine-tuning, for example, $Q(\mathcal{G}_L)+F(\mathcal{G}_U)$ means pretraining with graph $\mathcal{G}_L$ and fine-tune on graph $\mathcal{G}_U$.

First, it is worth noticing that $Q(\mathcal{G}_L)+F(\mathcal{G}_L)$ is the only variant that solely uses labeled data without inclusion of any unlabeled data in the training process. The results show that it performs the worst among all test cases. The takeaway here is that the using unlabeled data in the pretraining or fine-tuning process leads to a significant improvement in the model performance. Second, we compare the result of fine-tuning on $\mathcal{G}_L$ and $\mathcal{G}$. The results show that fine-tuning with unlabeled data generally performs better, which again shows the advantage of jointly considering labeled and unlabeled information. Lastly, we observe the difference between $Q(\mathcal{G}_L)$ , $Q(\mathcal{G}_U)$ and $Q(\mathcal{G})$. $Q(\mathcal{G}_L)$ is significantly worse than the others while $Q(\mathcal{G}_U)$ and $Q(\mathcal{G})$ usually gives similar performance. This final result adds to the already evident effectiveness of adding unlabeled data in the pretraining stage. It not only provides more training instances for learning the weight matrices (i.e., $\mathbf{W}_1^{(k)}, \mathbf{W}_2^{(k)}$) but makes the nearby transaction closer in embedding space, leading to better predictive capabilities.

\begin{figure}[t!]
\flushleft
\begin{minipage}[]{0.34\linewidth}
\centering
\includegraphics[width=0.8\linewidth]{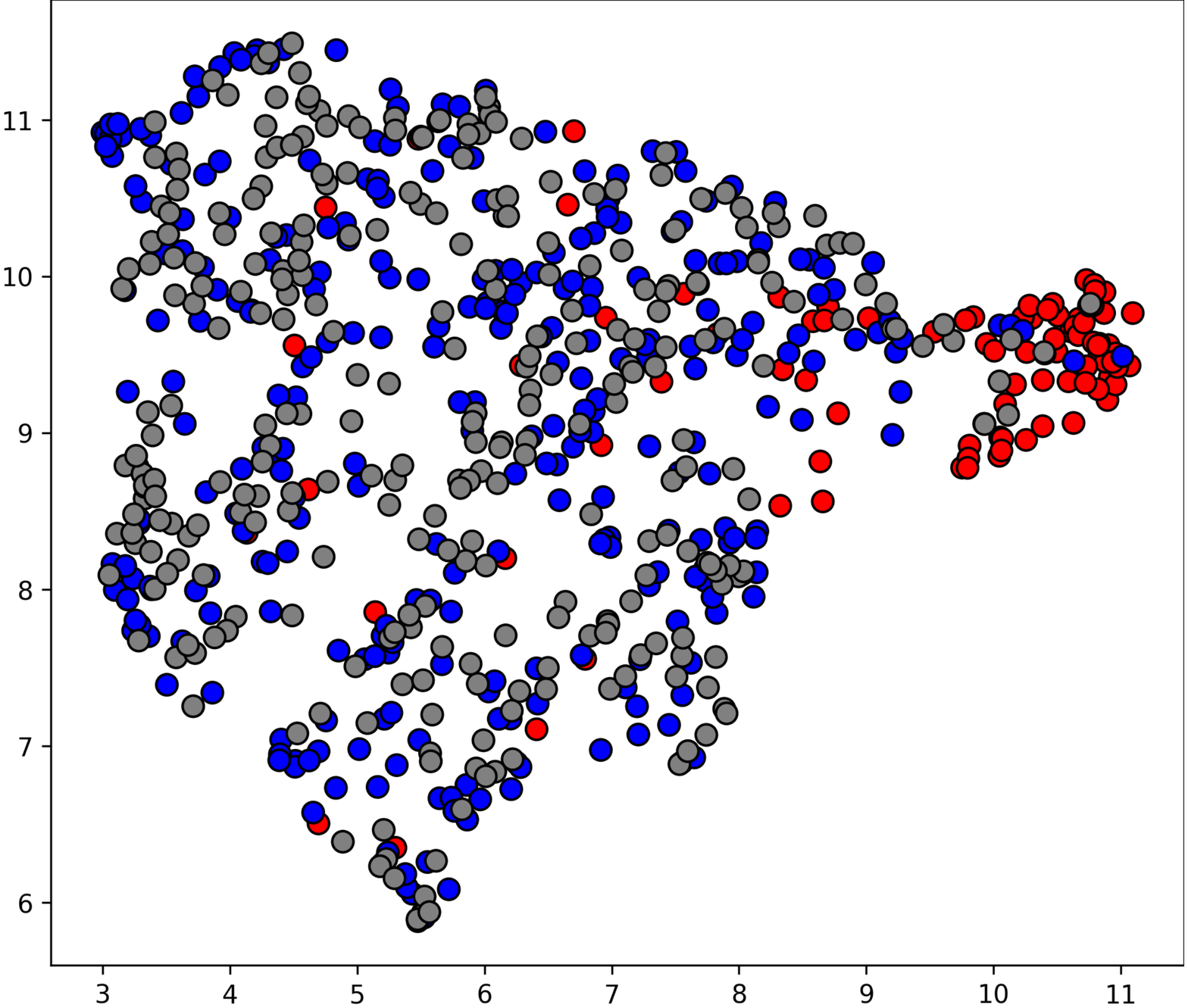}
\subcaption{w/o unlabeled data.}
\label{subfig:L_tsne}
\end{minipage}
\begin{minipage}[]{0.32\linewidth}
\centering
\includegraphics[width=0.85\linewidth]{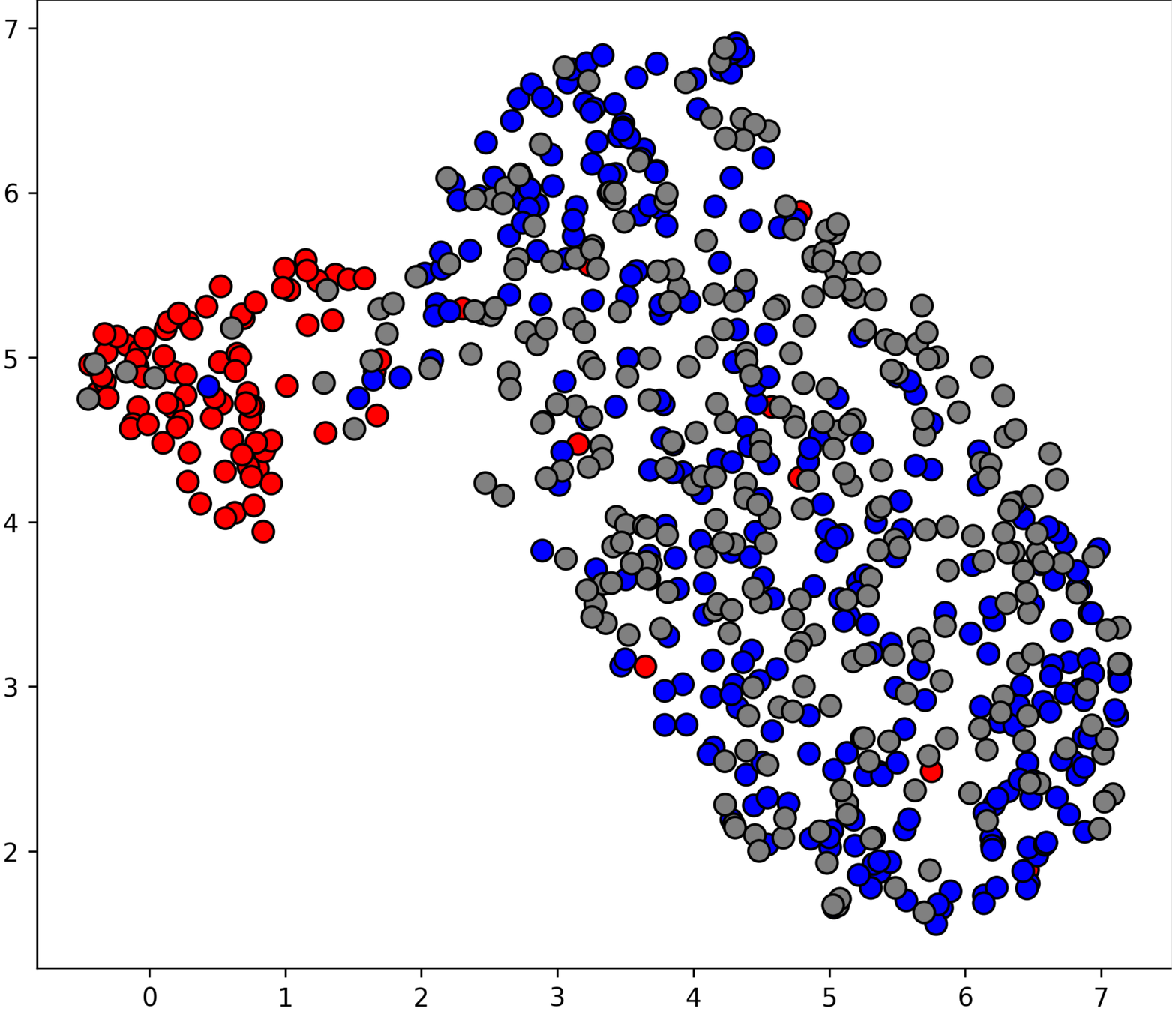}
\subcaption{w/o pretraining.}
\label{subfig:nopretrian_tsne}
\end{minipage}
\begin{minipage}[]{0.32\linewidth}
\centering
\includegraphics[width=0.9\linewidth]{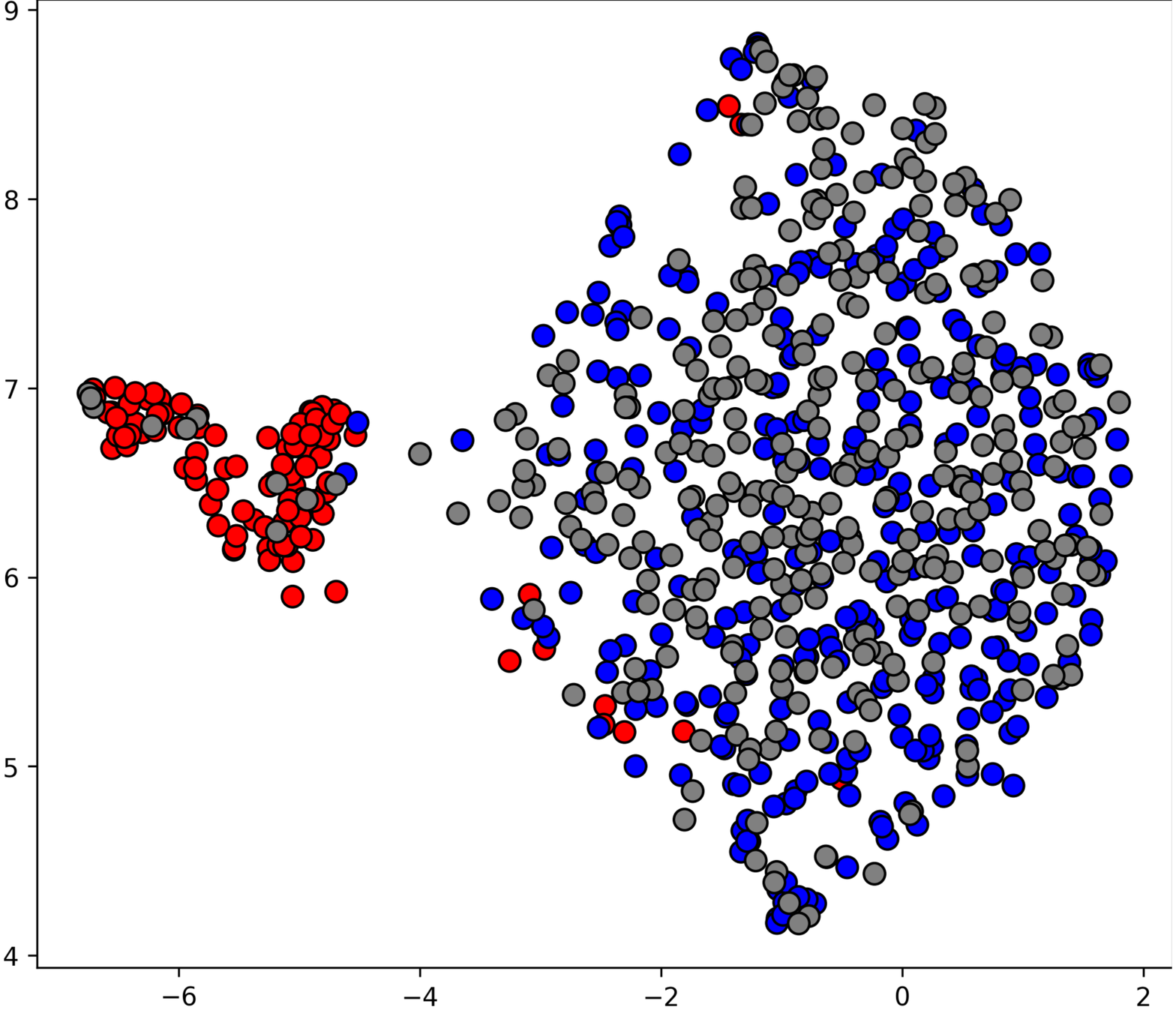}
\subcaption{Full model.}
\label{subfig:LU_tsne}
\end{minipage}
\vspace{-4mm}
\caption{TSNE visualization of variants of \model{}. Each transaction is mapped to a point in 2D space. Red, blue and grey dots denote the illicit, legitimate, and unlabeled transactions respectively.}
\label{fig:tsne_plot}
\end{figure}

To further explore the effect of including unlabeled data in \model{}, we train two variants of \model{} and mapped their transaction embedding obtainable before the output layer into 2-dimensional space using TSNE, and present the result in Figure~\ref{fig:tsne_plot}. Figure~\ref{subfig:L_tsne} presents the variant of \model{} where we remove all of the unlabeled data in $\mathcal{G}$ (refer to the setting of $Q(\mathcal{G}_L)+F(\mathcal{G}_L)$), whereas for~Figure~\ref{subfig:nopretrian_tsne}, the pretraining step is ommited, and Figure~\ref{subfig:LU_tsne} is the result of \model{}. Figure~\ref{subfig:LU_tsne} is the best performing case where
legitimate and illicit transactions are separated well into two tightly clustered sets. In Figure~\ref{subfig:L_tsne}, we can observe diverse patterns that are not clearly separable. Figure~\ref{subfig:nopretrian_tsne} has better separation but data points are more dispersed than as compared to Figure~\ref{subfig:LU_tsne}. To sum up, this analysis shows the importance of unlabeled data in learning high quality embedding vectors that played a key role for mitigating effects of label scarcity.

\subsection{Prediction Performance for Unseen Importers}
Inductive learning can be understood as the ability of the model to learn from the historic data and generalize it to new, unseen cases. 
Given the huge volume of transactions, the model will consistently come across input data that has unseen features, such as new items and importers. In the previous work DATE~\cite{kimtsai2020date}, the representations for every importer and HS-code were learnt, however, in a semi-supervised setting, such a learning becomes infeasible. \model{} provides a novel solution for inductive learning by linking the newly observed importer and item to the target transaction and its representation could be updated via multiple messages passing operation. 
To measure the degree of the inductive phenomenon, we introduce \emph{Out-of-Sample-Ratio} (OSR), which is defined as: 
$OSR = |P_{un}| / (|P_{s}| + |P_{un}|),$
where $|P_s|$ and $|P_{un}|$ denote the number of unique states (refer inductive setting in~\ref{lab:probsettin} for definition) observed and unobserved in the training data, respectively. The higher OSR is, the more unseen data states are present in the data. 
In this work, we focus on analyzing the behavior of importer and HS-code.

\begin{table}[ht!]
\caption{Inductive setting - Importer}
\label{tab:inductive_imp}
\small
\begin{tabular}{|c|c|c|c+c|c|c|c|}
\hline

\multicolumn{7}{|c|}{\textbf{A-Data}, OSR: 9.89\%}                \\ \hline
      & \multicolumn{3}{c+}{\textbf{n = 1\%}} & \multicolumn{3}{c|}{\textbf{n=5\%}} \\ \hline
Model & Pre.     & Rec.    & Rev.    & Pre.    & Rec.    & Rev.   \\ \hline
\model{}   & 0.0486   & 0.0168  & 0.0161  & 0.0396  & 0.0686  & 0.0672 \\ \hline
DATE  & 0.0660   & 0.0229  & 0.0271  & 0.0501  & 0.0866  & 0.0953 \\ \hline

\multicolumn{7}{|c|}{\textbf{B-Data}, OSR: 49.43\%}                 \\ \hline
\model{}   & \textbf{0.5386}   & \textbf{0.1963}  & \textbf{0.2236}  & \textbf{0.2055}  & \textbf{0.3744}  & \textbf{0.3846} \\ \hline
DATE  & 0.0655   & 0.0239  & 0.0482  & 0.0549  & 0.1000  & 0.1580 \\ \hline

\multicolumn{7}{|c|}{\textbf{C-Data}, OSR: 5.09\%}                 \\ \hline
\model{}   & \textbf{0.9052}   & \textbf{0.0483}  & \textbf{0.0847}  & \textbf{0.7805}  & \textbf{0.2081}  & \textbf{0.3396} \\ \hline
DATE  & 0.7730   & 0.0413  & 0.0616  & 0.6418  & 0.1711  & 0.2605 \\ \hline

\end{tabular}
\end{table}

\begin{table}[ht!]
\caption{Inductive setting - HSCode}
\label{tab:inductive_hs}
\small
\begin{tabular}{|c|c|c|c+c|c|c|c|}
\hline

\multicolumn{7}{|c|}{\textbf{A-Data}, OSR: 2.75\%}                \\ \hline
      & \multicolumn{3}{c+}{\textbf{n = 1\%}} & \multicolumn{3}{c|}{\textbf{n=5\%}} \\ \hline
Model & Pre.     & Rec.    & Rev.    & Pre.    & Rec.    & Rev.   \\ \hline
\model{}   & \textbf{0.0560}   & \textbf{0.0329}  & \textbf{0.0219}  & \textbf{0.0471}  & \textbf{0.1382}  & \textbf{0.1213} \\ \hline
DATE  & 0.0373   & 0.0219  & 0.0104  & 0.0396  & 0.1162  & 0.0964 \\ \hline

\multicolumn{7}{|c|}{\textbf{B-Data}, OSR: 3.64\%}                 \\ \hline
\model{}   & \textbf{0.4554}   & \textbf{0.1053}  & \textbf{0.1256}  & \textbf{0.1577}  & \textbf{0.1808}  & 0.1889 \\ \hline
DATE  & 0.2772   & 0.0641  & 0.1002  & 0.0978  & 0.1121  & \textbf{0.1960} \\ \hline

\multicolumn{7}{|c|}{\textbf{C-Data}, OSR: 4.74\%}                 \\ \hline
\model{}   & \textbf{0.7320}   & \textbf{0.0574}  & \textbf{0.1143}  & \textbf{0.4277}  & \textbf{0.1675}  & \textbf{0.2620} \\ \hline
DATE  & 0.5876   & 0.0461  & 0.0933  & 0.3946  & 0.1545  & 0.2142 \\ \hline

\end{tabular}
\end{table}

\begin{figure*}
\caption{Performance comparison by varying OSR (\%). (a) \& (b) \& (c) (controls importer id), (d) \& (e) \& (f) (controls HS-code). In all cases, \model{} outperforms the baselines.
\label{fig:varying_osr}
}
\begin{minipage}[t!]{0.33\linewidth}
\centering
\includegraphics[width=2.3in]{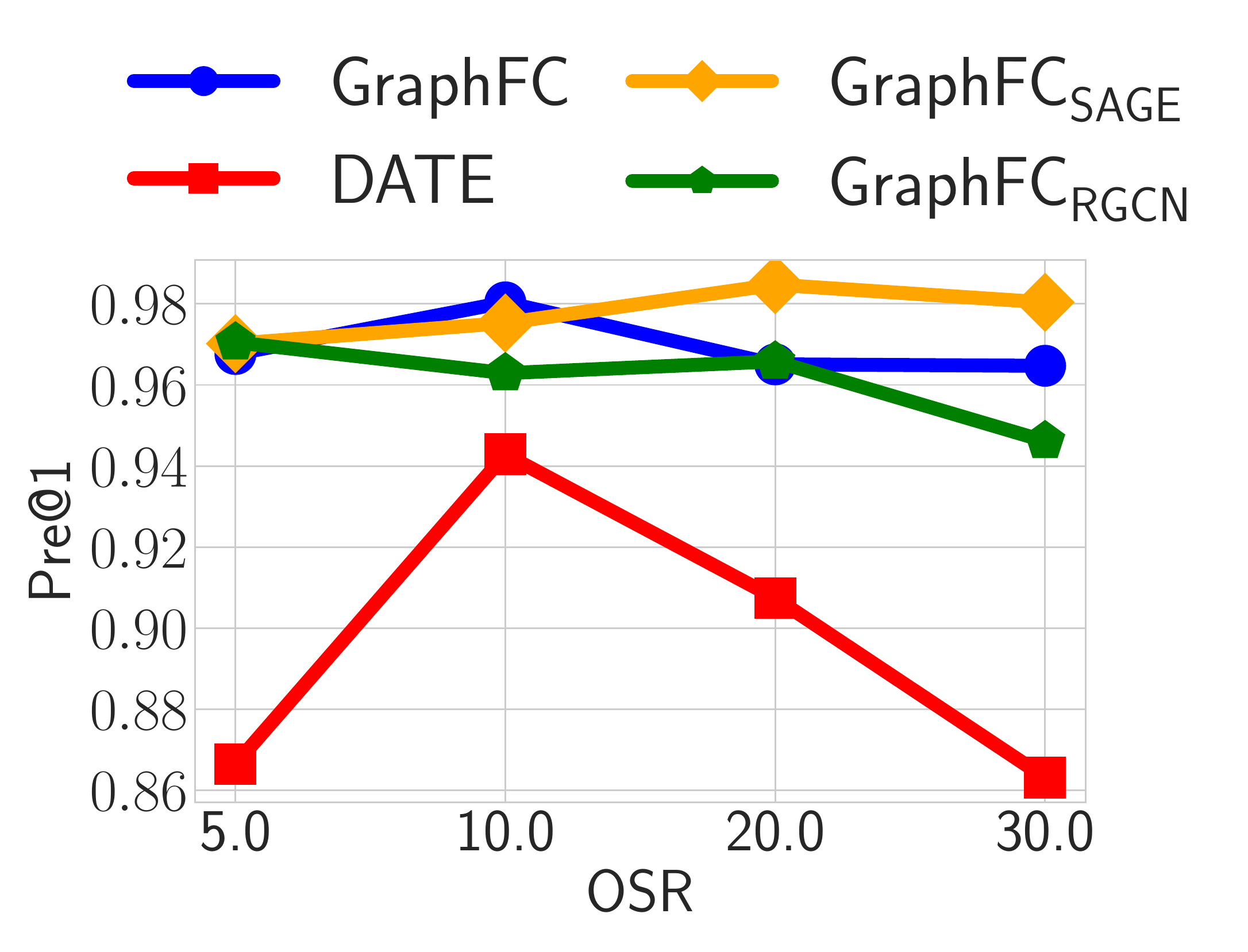}
\subcaption{N-data}
\end{minipage}%
\begin{minipage}[t!]{0.33\linewidth}
\centering
\includegraphics[width=2.3in]{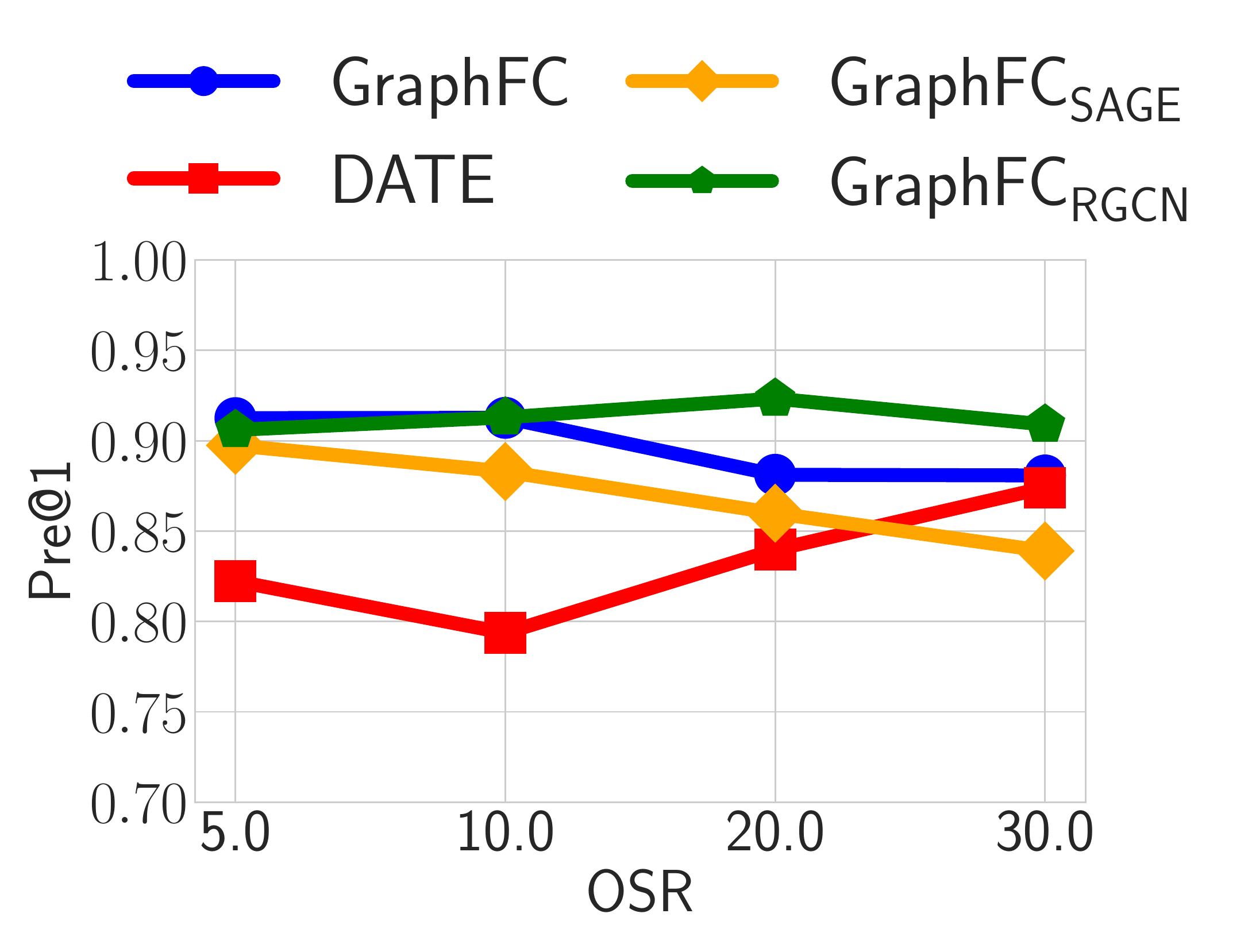}
\subcaption{T-data}
\end{minipage}
\begin{minipage}[t!]{0.33\linewidth}
\centering
\includegraphics[width=2.3in]{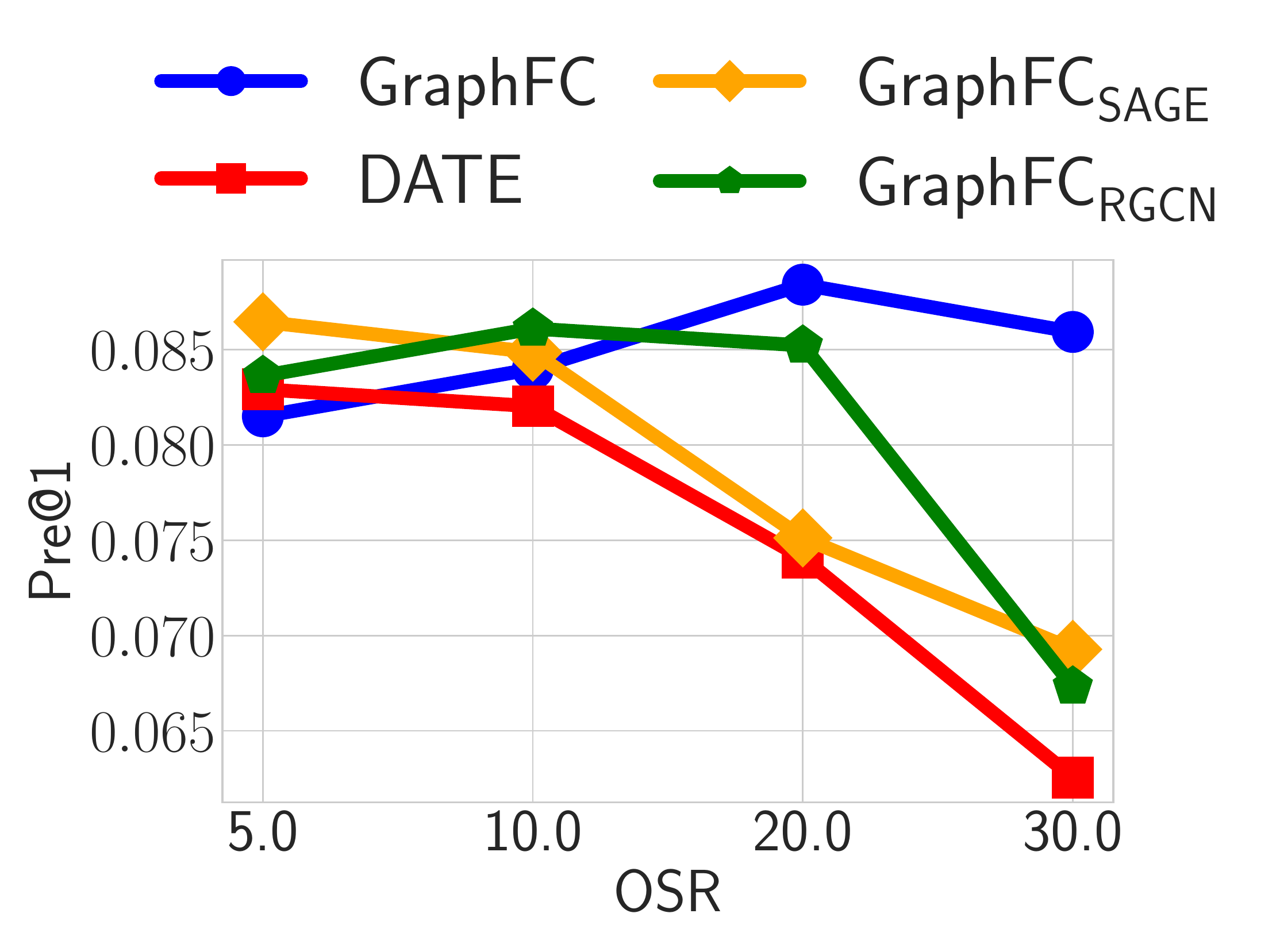}
\subcaption{C-data}
\end{minipage}
\begin{minipage}[t!]{0.33\linewidth}
\centering
\includegraphics[width=2.3in]{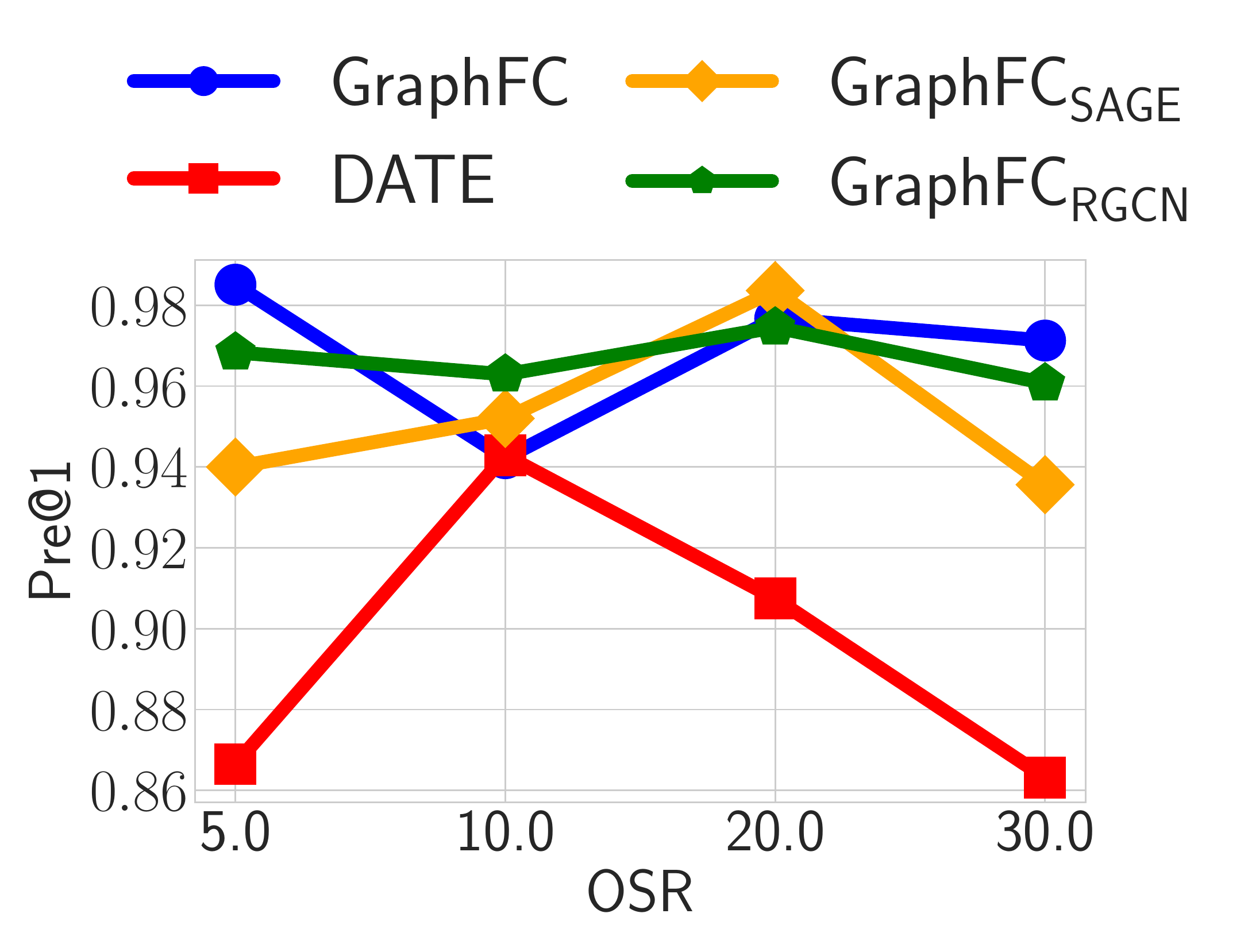}
\subcaption{N-data}
\end{minipage}%
\begin{minipage}[t!]{0.33\linewidth}
\centering
\includegraphics[width=2.3in]{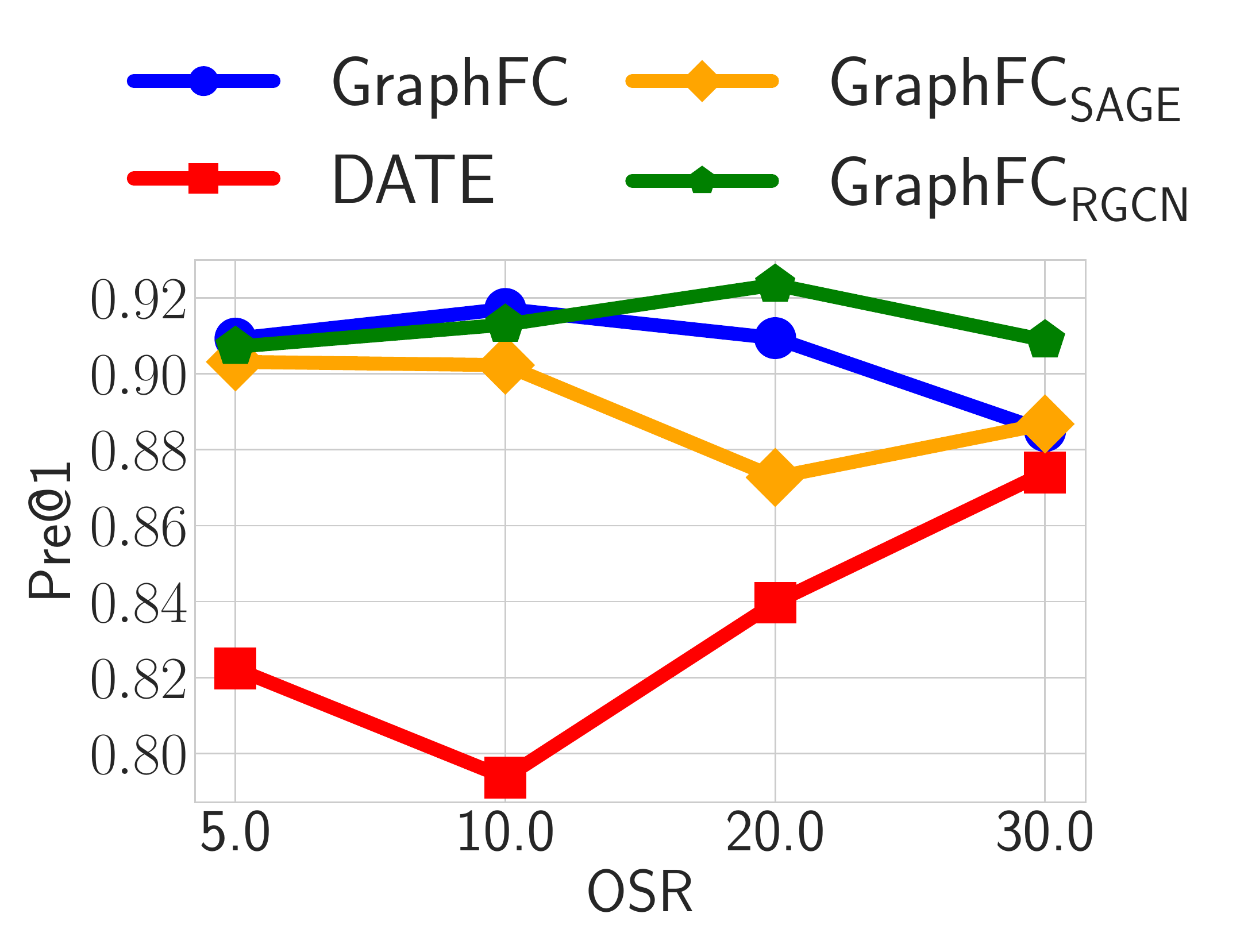}
\subcaption{T-data}
\end{minipage}
\begin{minipage}[t!]{0.33\linewidth}
\centering

\includegraphics[width=2.3in]{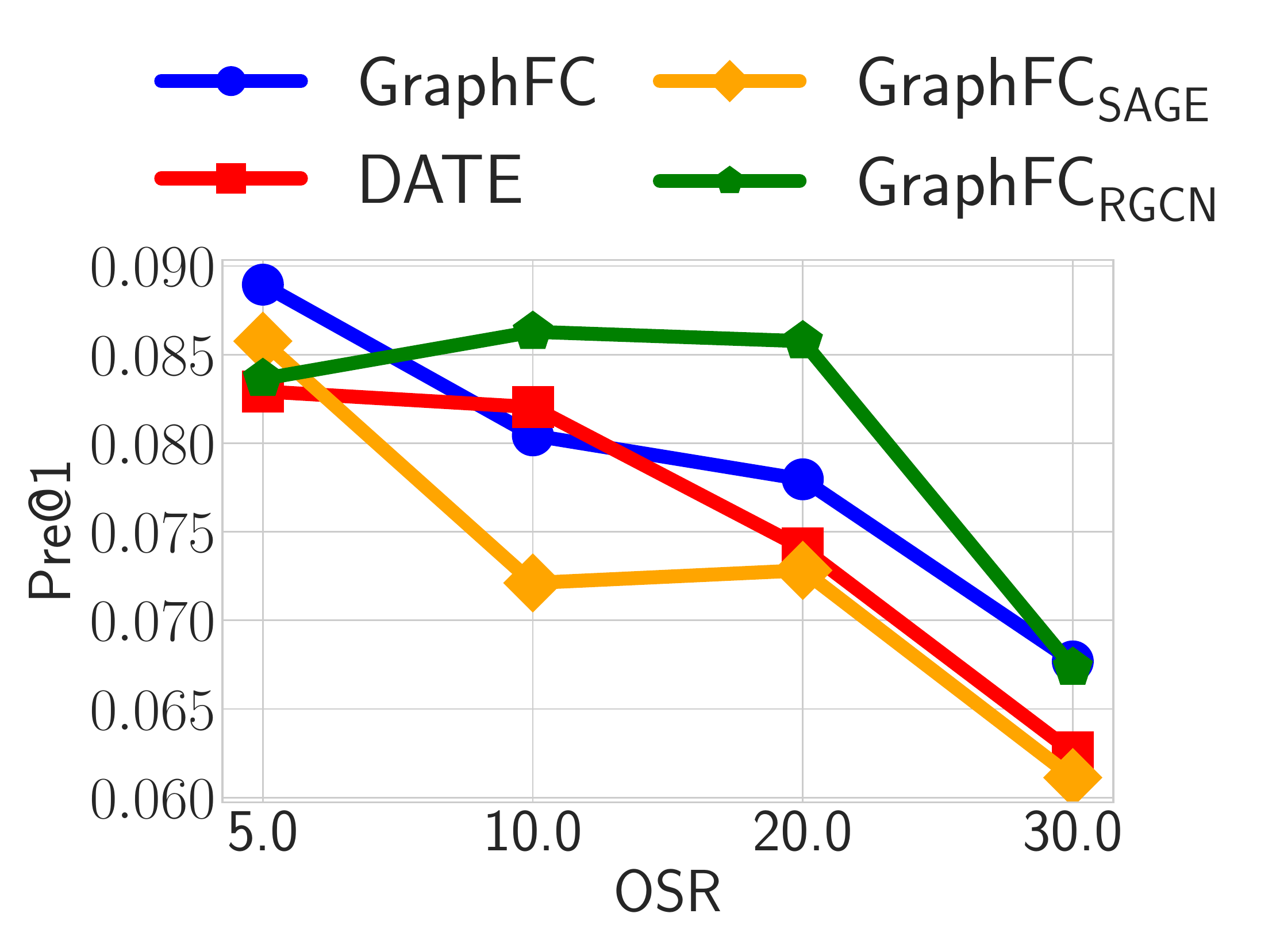}
\subcaption{C-data}
\end{minipage}
\end{figure*}

\textbf{Inductive analysis in semi-supervised learning}
We evaluate the performance of inductive learning by selecting a subset of transactions from testing data where the importer ID or HS-code were not observed in training data. The result is presented in
and presents the result in
Table~\ref{tab:inductive_imp} and~\ref{tab:inductive_hs}. We compare \model{} with DATE as DATE being the state-of-the-art method for transductive learning.
In Table~\ref{tab:inductive_imp}, it demonstrates the performance on predicting transactions made by unseen importers. \model{} outperforms DATE with a significant improvement as OSR increases. Especially in B-data, the OSR is 49.43\% and \model{} gains 7x improvement in terms of Pre@1\%. On the other hand, Table~\ref{tab:inductive_hs} also shows the superiority against DATE on unseen HS-code. Additionally, in Figure~\ref{fig:varying_osr} we also show the effect of varying OSR rates by controlling HS codes and importer id. \model{} outperforms the baselines in all cases.

\subsection{Component Analysis}

\begin{table}[t!]
\caption{Component Analysis.}
\label{tab:ablation}
\small
\centering
\begin{tabular}{|c|c|c|c|c|c|c|}
\hline
\multicolumn{7}{|c|}{\textbf{A-Data}} \\ \hline
& \multicolumn{3}{c|}{\textbf{n = 1\%}} & \multicolumn{3}{c|}{\textbf{n=5\%}} \\ \hline
Model & Pre. & Rec. & Rev. & Pre. & Rec. & Rev.   \\ \hline

\model{}\textsubscript{semi}  & 0.030 & 0.024 & 0.034  & 0.023 & 0.092 & 0.127 \\ \hline
\model{}\textsubscript{joint}  & 0.025 & 0.019 & 0.037 & 0.023 & 0.090 & 0.117 \\ \hline
\model{}\textsubscript{only} & 0.032 & 0.024 & 0.034& 0.017 & 0.067 & 0.081 \\ \hline
\model{}\textsubscript{sparse} & 0.030 & 0.024 & 0.037& 0.018 & 0.070 & 0.096 \\ \hline
\model{} & \textbf{0.045} & \textbf{0.035} & \textbf{0.057}& \textbf{0.030} & \textbf{0.118} & \textbf{0.170} \\ \hline
\multicolumn{7}{|c|}{\textbf{B-Data}} \\ \hline
\model{}\textsubscript{semi}  & 0.281 & 0.113 & 0.162  & 0.197 & 0.398 & \textbf{0.367}  \\ \hline
\model{}\textsubscript{joint}  & 0.218 & 0.076 & 0.092 & 0.105 & 0.203 & 0.212 \\ \hline
\model{}\textsubscript{only} & 0.068 & 0.027 & 0.069 & 0.034 & 0.069 & 0.157 \\ \hline
\model{}\textsubscript{sparse} & 0.067 & 0.070 & 0.120& 0.044 & 0.234 & 0.359 \\ \hline
\model{} & \textbf{0.401} & \textbf{0.162} & \textbf{0.171} & \textbf{0.200} & \textbf{0.405} & 0.304 \\ \hline
\multicolumn{7}{|c|}{\textbf{C-Data}} \\ \hline
\model{}\textsubscript{semi}  & 0.829 & 0.087 & 0.143 & \textbf{0.548} & \textbf{0.289} & 0.434 \\ \hline
\model{}\textsubscript{joint}  & 0.807 & 0.061 & 0.135& 0.491 & 0.249 & 0.401\\ \hline
\model{}\textsubscript{only} & 0.764 & 0.080 & 0.115 & 0.490 & 0.258 & 0.382 \\ \hline
\model{}\textsubscript{sparse} & 0.712 & 0.075 & 0.123& 0.473 & 0.250 & 0.376\\ \hline
\model{} & \textbf{0.869} & \textbf{0.092} & \textbf{0.170} & 0.535 & 0.282 & \textbf{0.445} \\ \hline
\end{tabular}
\end{table}
To establish the contribution of different components in~\model{}, we ablate its core components and evaluate its performance. Specifically, we evaluate the following settings:

\begin{itemize}[leftmargin=*]
    \item \textbf{GraphFC}: Use full model.
    \item \textbf{No unsupervised pretraining (GNN\textsubscript{semi})}: Skip the unsupervised pre-train step.
    \item \textbf{Joint training (GNN\textsubscript{joint})}: Remove pretraining and jointly optimizing E.q~\ref{loss:pretrain} and Eq.~\ref{eq:loss}.
    \item \textbf{No GBDT (GNN\textsubscript{only})}: Remove the GBDT step of the model and utilize the data from transactions directly.
    \item \textbf{No item/importer nodes (GNN\textsubscript{sparse})}: Remove one of the categorical nodes utilized for building the graph structure and build a \textit{sparser} graph. 
\end{itemize}

Results of different variants are demonstrate in Table~\ref{tab:ablation}. 
It can be noticed that removing any component in \model{} leads to degradation of performance. The only exception is for $\model{}\textsubscript{semi}$  that delivers similar performance in C-data. Among these variants, $\model{}\textsubscript{sparse}$ and $\model{}\textsubscript{only}$ perform poorly which indicates that both the GBDT step, and utilizing the importer and HS-code for constructing the transaction graph play an important role in \model{}. 
 
In general, neither $\model{}\textsubscript{semi}$ nor $\model{}\textsubscript{joint}$ improves \model{} which verifies the necessity our pretraining step. Specifically, the difference between \model{} and $\model{}\textsubscript{semi}$, $\model{}\textsubscript{joint}$ are significant when $n$ is small (e.g. $n=1\%$) which is important since customs usually maintain a low inpsection rate lower than 5\%.

\section{Discussion and Conclusion}

With increased connectivity, development, and globalization, customs administrations across the world have to process an increasingly large number of transactions. This calls for an automated system for detecting fraudulent transactions, as manual inspection of such an astronomical volume of transactions is infeasible. In this work, we develop~\model{}, a GNN based fraud detection model that transforms the customs tabular data into a graph and maximizes the identification of malicious customs transactions while also maximizing the additional revenue collected from these transactions. We propose the given approach and exhibit its effectiveness by developing a model and testing it on real data from multiple countries. In our experiments, we show that \model{} is capable of making use of the precious unlabeled data and further substantially alleviate the performance drop under different levels of label scarcity. The proposed model provides a remedy to any country who is trying to facilitate automated customs fraud detection even if the administration did not collect huge labeled data.

\bibliographystyle{ACM-Reference-Format}
\bibliography{main}


\begin{thebibliography}{48}


\ifx \showCODEN    \undefined \def \showCODEN     #1{\unskip}     \fi
\ifx \showDOI      \undefined \def \showDOI       #1{#1}\fi
\ifx \showISBNx    \undefined \def \showISBNx     #1{\unskip}     \fi
\ifx \showISBNxiii \undefined \def \showISBNxiii  #1{\unskip}     \fi
\ifx \showISSN     \undefined \def \showISSN      #1{\unskip}     \fi
\ifx \showLCCN     \undefined \def \showLCCN      #1{\unskip}     \fi
\ifx \shownote     \undefined \def \shownote      #1{#1}          \fi
\ifx \showarticletitle \undefined \def \showarticletitle #1{#1}   \fi
\ifx \showURL      \undefined \def \showURL       {\relax}        \fi
\providecommand\bibfield[2]{#2}
\providecommand\bibinfo[2]{#2}
\providecommand\natexlab[1]{#1}
\providecommand\showeprint[2][]{arXiv:#2}

\bibitem[\protect\citeauthoryear{Abdallah, Maarof, and Zainal}{Abdallah
  et~al\mbox{.}}{2016}]%
        {ABDALLAH201690}
\bibfield{author}{\bibinfo{person}{Aisha Abdallah},
  \bibinfo{person}{Mohd~Aizaini Maarof}, {and} \bibinfo{person}{Anazida
  Zainal}.} \bibinfo{year}{2016}\natexlab{}.
\newblock \showarticletitle{Fraud detection system: A survey}.
\newblock \bibinfo{journal}{\emph{Journal of Network and Computer
  Applications}}  \bibinfo{volume}{68} (\bibinfo{year}{2016}).
\newblock
\showISSN{1084--8045}


\bibitem[\protect\citeauthoryear{Adewumi and Akinyelu}{Adewumi and
  Akinyelu}{2017}]%
        {adewumi2017survey}
\bibfield{author}{\bibinfo{person}{Aderemi~O Adewumi} {and}
  \bibinfo{person}{Andronicus~A Akinyelu}.} \bibinfo{year}{2017}\natexlab{}.
\newblock \showarticletitle{A survey of machine-learning and nature-inspired
  based credit card fraud detection techniques}.
\newblock \bibinfo{journal}{\emph{International Journal of System Assurance
  Engineering and Management}} \bibinfo{volume}{8}, \bibinfo{number}{2}
  (\bibinfo{year}{2017}).
\newblock


\bibitem[\protect\citeauthoryear{Ahmed, Mahmood, and Islam}{Ahmed
  et~al\mbox{.}}{2016}]%
        {AHMED2016278}
\bibfield{author}{\bibinfo{person}{Mohiuddin Ahmed},
  \bibinfo{person}{Abdun~Naser Mahmood}, {and} \bibinfo{person}{Md.~Rafiqul
  Islam}.} \bibinfo{year}{2016}\natexlab{}.
\newblock \showarticletitle{A survey of anomaly detection techniques in
  financial domain}.
\newblock \bibinfo{journal}{\emph{Future Generation Computer Systems}}
  \bibinfo{volume}{55} (\bibinfo{year}{2016}), \bibinfo{pages}{278--288}.
\newblock


\bibitem[\protect\citeauthoryear{Arik and Pfister}{Arik and Pfister}{2019}]%
        {arik2019tabnet}
\bibfield{author}{\bibinfo{person}{Sercan~O Arik} {and} \bibinfo{person}{Tomas
  Pfister}.} \bibinfo{year}{2019}\natexlab{}.
\newblock \showarticletitle{Tabnet: Attentive interpretable tabular learning}.
\newblock \bibinfo{journal}{\emph{arXiv preprint arXiv:1908.07442}}
  (\bibinfo{year}{2019}).
\newblock


\bibitem[\protect\citeauthoryear{Arvis, Ojala, Wiederer, Shepherd, Raj,
  Dairabayeva, and Kiiski}{Arvis et~al\mbox{.}}{2018}]%
        {worldbank}
\bibfield{author}{\bibinfo{person}{Jean-Fran{\c{c}}ois Arvis},
  \bibinfo{person}{Lauri Ojala}, \bibinfo{person}{Christina Wiederer},
  \bibinfo{person}{Ben Shepherd}, \bibinfo{person}{Anasuya Raj},
  \bibinfo{person}{Karlygash Dairabayeva}, {and} \bibinfo{person}{Tuomas
  Kiiski}.} \bibinfo{year}{2018}\natexlab{}.
\newblock \showarticletitle{Connecting to compete 2018}.
\newblock  (\bibinfo{year}{2018}).
\newblock


\bibitem[\protect\citeauthoryear{Bojchevski and Günnemann}{Bojchevski and
  Günnemann}{2018}]%
        {g2g}
\bibfield{author}{\bibinfo{person}{Aleksandar Bojchevski} {and}
  \bibinfo{person}{Stephan Günnemann}.} \bibinfo{year}{2018}\natexlab{}.
\newblock \bibinfo{title}{Deep Gaussian Embedding of Graphs: Unsupervised
  Inductive Learning via Ranking}.
\newblock
\newblock
\showeprint[arxiv]{1707.03815}~[stat.ML]


\bibitem[\protect\citeauthoryear{Canrakerta, Hidayanto, and
  Ruldeviyani}{Canrakerta et~al\mbox{.}}{2020}]%
        {Canrakerta_2020}
\bibfield{author}{\bibinfo{person}{Canrakerta}, \bibinfo{person}{Achmad~Nizar
  Hidayanto}, {and} \bibinfo{person}{Yova Ruldeviyani}.}
  \bibinfo{year}{2020}\natexlab{}.
\newblock \showarticletitle{Application of business intelligence for customs
  declaration: A case study in Indonesia}.
\newblock \bibinfo{journal}{\emph{Journal of Physics: Conference Series}}
  \bibinfo{volume}{1444} (\bibinfo{year}{2020}), \bibinfo{pages}{012028}.
\newblock


\bibitem[\protect\citeauthoryear{Cerioli, Barabesi, Cerasa, Menegatti, and
  Perrotta}{Cerioli et~al\mbox{.}}{2019}]%
        {web3}
\bibfield{author}{\bibinfo{person}{Andrea Cerioli}, \bibinfo{person}{Lucio
  Barabesi}, \bibinfo{person}{Andrea Cerasa}, \bibinfo{person}{Mario
  Menegatti}, {and} \bibinfo{person}{Domenico Perrotta}.}
  \bibinfo{year}{2019}\natexlab{}.
\newblock \showarticletitle{Newcomb{\textendash}Benford law and the detection
  of frauds in international trade}.
\newblock \bibinfo{journal}{\emph{Proceedings of the National Academy of
  Sciences}} \bibinfo{volume}{116}, \bibinfo{number}{1} (\bibinfo{year}{2019}),
  \bibinfo{pages}{106--115}.
\newblock


\bibitem[\protect\citeauthoryear{Chen and Guestrin}{Chen and Guestrin}{2016}]%
        {tianqi2016}
\bibfield{author}{\bibinfo{person}{Tianqi Chen} {and} \bibinfo{person}{Carlos
  Guestrin}.} \bibinfo{year}{2016}\natexlab{}.
\newblock \showarticletitle{{XGBoost: A scalable tree boosting system}}. In
  \bibinfo{booktitle}{\emph{KDD}}. \bibinfo{pages}{785--794}.
\newblock


\bibitem[\protect\citeauthoryear{Chen and Sun}{Chen and Sun}{2020}]%
        {gnnanomaly}
\bibfield{author}{\bibinfo{person}{Zhe Chen} {and} \bibinfo{person}{Aixin
  Sun}.} \bibinfo{year}{2020}\natexlab{}.
\newblock \showarticletitle{Anomaly Detection on Dynamic Bipartite Graph with
  Burstiness}. In \bibinfo{booktitle}{\emph{2020 IEEE International Conference
  on Data Mining (ICDM)}}. \bibinfo{pages}{966--971}.
\newblock


\bibitem[\protect\citeauthoryear{de~Roux, Perez, Moreno, Villamil, and
  Figueroa}{de~Roux et~al\mbox{.}}{2018}]%
        {deRoux2018taxkdd}
\bibfield{author}{\bibinfo{person}{Daniel de Roux}, \bibinfo{person}{Boris
  Perez}, \bibinfo{person}{Andr{\'e}s Moreno}, \bibinfo{person}{Maria del~Pilar
  Villamil}, {and} \bibinfo{person}{C{\'e}sar Figueroa}.}
  \bibinfo{year}{2018}\natexlab{}.
\newblock \showarticletitle{Tax fraud detection for under-reporting
  declarations using an unsupervised machine learning approach}. In
  \bibinfo{booktitle}{\emph{KDD}}. \bibinfo{pages}{215--222}.
\newblock


\bibitem[\protect\citeauthoryear{Dou, Liu, Sun, Deng, Peng, and Yu}{Dou
  et~al\mbox{.}}{2020}]%
        {gnncamouflaged2020}
\bibfield{author}{\bibinfo{person}{Yingtong Dou}, \bibinfo{person}{Zhiwei Liu},
  \bibinfo{person}{Li Sun}, \bibinfo{person}{Yutong Deng}, \bibinfo{person}{Hao
  Peng}, {and} \bibinfo{person}{Philip~S. Yu}.}
  \bibinfo{year}{2020}\natexlab{}.
\newblock \showarticletitle{Enhancing Graph Neural Network-Based Fraud
  Detectors against Camouflaged Fraudsters}. In
  \bibinfo{booktitle}{\emph{Proceedings of the 29th ACM International
  Conference on Information \& Knowledge Management}}.
  \bibinfo{pages}{315–324}.
\newblock


\bibitem[\protect\citeauthoryear{Filho and Wainer}{Filho and Wainer}{2008}]%
        {jambeiro08jmlr}
\bibfield{author}{\bibinfo{person}{Jorge~Jambeiro Filho} {and}
  \bibinfo{person}{Jacques Wainer}.} \bibinfo{year}{2008}\natexlab{}.
\newblock \showarticletitle{{HPB: A model for handling BN nodes with high
  cardinality parents}}.
\newblock \bibinfo{journal}{\emph{JMLR}}  \bibinfo{volume}{9}
  (\bibinfo{year}{2008}), \bibinfo{pages}{2141--2170}.
\newblock


\bibitem[\protect\citeauthoryear{Geourjon, Laporte, Coundoul, Gadiaga, Cantens,
  Ireland, and Raballand}{Geourjon et~al\mbox{.}}{2013}]%
        {geourjon2013inspecting}
\bibfield{author}{\bibinfo{person}{Anne-Marie Geourjon},
  \bibinfo{person}{Bertrand Laporte}, \bibinfo{person}{Ousmane Coundoul},
  \bibinfo{person}{Massene Gadiaga}, \bibinfo{person}{T Cantens},
  \bibinfo{person}{R Ireland}, {and} \bibinfo{person}{G Raballand}.}
  \bibinfo{year}{2013}\natexlab{}.
\newblock \showarticletitle{Inspecting Less to Inspect Better. The Use of Data
  Mining for Risk Management by Customs Administrations'}.
\newblock \bibinfo{journal}{\emph{Reform by Numbers. Measurement Applied to
  Customs and Tax Administrations in Developing Countries, Washington DC: World
  Bank}} (\bibinfo{year}{2013}).
\newblock


\bibitem[\protect\citeauthoryear{Grover and Leskovec}{Grover and
  Leskovec}{2016}]%
        {grover2016node2vec}
\bibfield{author}{\bibinfo{person}{Aditya Grover} {and} \bibinfo{person}{Jure
  Leskovec}.} \bibinfo{year}{2016}\natexlab{}.
\newblock \showarticletitle{node2vec: Scalable feature learning for networks}.
  In \bibinfo{booktitle}{\emph{Proceedings of the 22nd ACM SIGKDD international
  conference on Knowledge discovery and data mining}}.
  \bibinfo{pages}{855--864}.
\newblock


\bibitem[\protect\citeauthoryear{Hamilton, Ying, and Leskovec}{Hamilton
  et~al\mbox{.}}{2017a}]%
        {graphsage}
\bibfield{author}{\bibinfo{person}{Will Hamilton}, \bibinfo{person}{Zhitao
  Ying}, {and} \bibinfo{person}{Jure Leskovec}.}
  \bibinfo{year}{2017}\natexlab{a}.
\newblock \showarticletitle{Inductive representation learning on large graphs}.
  In \bibinfo{booktitle}{\emph{proc. of the NeurIPS}}.
  \bibinfo{pages}{1024--1034}.
\newblock


\bibitem[\protect\citeauthoryear{Hamilton, Ying, and Leskovec}{Hamilton
  et~al\mbox{.}}{2017b}]%
        {hamilton2017inductive}
\bibfield{author}{\bibinfo{person}{William~L Hamilton}, \bibinfo{person}{Rex
  Ying}, {and} \bibinfo{person}{Jure Leskovec}.}
  \bibinfo{year}{2017}\natexlab{b}.
\newblock \showarticletitle{Inductive representation learning on large graphs}.
\newblock \bibinfo{journal}{\emph{arXiv preprint arXiv:1706.02216}}
  (\bibinfo{year}{2017}).
\newblock


\bibitem[\protect\citeauthoryear{He, Pan, Jin, Xu, Liu, Xu, Shi, Atallah,
  Herbrich, Bowers, et~al\mbox{.}}{He et~al\mbox{.}}{2014}]%
        {he2014adkdd}
\bibfield{author}{\bibinfo{person}{Xinran He}, \bibinfo{person}{Junfeng Pan},
  \bibinfo{person}{Ou Jin}, \bibinfo{person}{Tianbing Xu}, \bibinfo{person}{Bo
  Liu}, \bibinfo{person}{Tao Xu}, \bibinfo{person}{Yanxin Shi},
  \bibinfo{person}{Antoine Atallah}, \bibinfo{person}{Ralf Herbrich},
  \bibinfo{person}{Stuart Bowers}, {et~al\mbox{.}}}
  \bibinfo{year}{2014}\natexlab{}.
\newblock \showarticletitle{Practical lessons from predicting clicks on ads at
  facebook}. In \bibinfo{booktitle}{\emph{ADKDD}}. \bibinfo{pages}{1--9}.
\newblock


\bibitem[\protect\citeauthoryear{Keen}{Keen}{2003}]%
        {web2}
\bibfield{author}{\bibinfo{person}{Michael Keen}.}
  \bibinfo{year}{2003}\natexlab{}.
\newblock \bibinfo{booktitle}{\emph{Changing Customs: Challenges and Strategies
  for the Reform of Customs Administration}}.
\newblock \bibinfo{publisher}{International Monetary Fund},
  \bibinfo{address}{USA}.
\newblock
\showISBNx{9781589062115}


\bibitem[\protect\citeauthoryear{Kim, Tsai, Singh, Choi, Ibok, Li, and Cha}{Kim
  et~al\mbox{.}}{2020}]%
        {kimtsai2020date}
\bibfield{author}{\bibinfo{person}{Sundong Kim}, \bibinfo{person}{Yu-Che Tsai},
  \bibinfo{person}{Karandeep Singh}, \bibinfo{person}{Yeonsoo Choi},
  \bibinfo{person}{Etim Ibok}, \bibinfo{person}{Cheng-Te Li}, {and}
  \bibinfo{person}{Meeyoung Cha}.} \bibinfo{year}{2020}\natexlab{}.
\newblock \showarticletitle{DATE: Dual Attentive Tree-aware Embedding for
  Customs Fraud Detection}. In \bibinfo{booktitle}{\emph{Proceedings of the
  26th ACM SIGKDD International Conference on Knowledge Discovery and Data
  Mining}}.
\newblock


\bibitem[\protect\citeauthoryear{Kipf and Welling}{Kipf and Welling}{2016}]%
        {kipf2016variational}
\bibfield{author}{\bibinfo{person}{Thomas~N. Kipf} {and} \bibinfo{person}{Max
  Welling}.} \bibinfo{year}{2016}\natexlab{}.
\newblock \bibinfo{title}{Variational Graph Auto-Encoders}.
\newblock
\newblock
\showeprint[arxiv]{1611.07308}~[stat.ML]


\bibitem[\protect\citeauthoryear{Kipf and Welling}{Kipf and Welling}{2017}]%
        {gcn}
\bibfield{author}{\bibinfo{person}{Thomas~N. Kipf} {and} \bibinfo{person}{Max
  Welling}.} \bibinfo{year}{2017}\natexlab{}.
\newblock \showarticletitle{Semi-Supervised Classification with Graph
  Convolutional Networks}. In \bibinfo{booktitle}{\emph{Proc. of the ICLR}}.
\newblock


\bibitem[\protect\citeauthoryear{Krivko}{Krivko}{2010}]%
        {KRIVKO20106070}
\bibfield{author}{\bibinfo{person}{Maria Krivko}.}
  \bibinfo{year}{2010}\natexlab{}.
\newblock \showarticletitle{A hybrid model for plastic card fraud detection
  systems}.
\newblock \bibinfo{journal}{\emph{Expert Systems with Applications}}
  \bibinfo{volume}{37}, \bibinfo{number}{8} (\bibinfo{year}{2010}),
  \bibinfo{pages}{6070--6076}.
\newblock


\bibitem[\protect\citeauthoryear{Kumar and Nagadevara}{Kumar and
  Nagadevara}{2006}]%
        {kumar2006development}
\bibfield{author}{\bibinfo{person}{Anuj Kumar} {and}
  \bibinfo{person}{Vishnuprasad Nagadevara}.} \bibinfo{year}{2006}\natexlab{}.
\newblock \showarticletitle{Development of hybrid classification methodology
  for mining skewed data sets: A case study of Indian customs data}.
\newblock  (\bibinfo{year}{2006}).
\newblock


\bibitem[\protect\citeauthoryear{Kültür and Çağlayan}{Kültür and
  Çağlayan}{2017}]%
        {Kultur}
\bibfield{author}{\bibinfo{person}{Yiğit Kültür} {and}
  \bibinfo{person}{Mehmet~Ufuk Çağlayan}.} \bibinfo{year}{2017}\natexlab{}.
\newblock \showarticletitle{Hybrid approaches for detecting credit card fraud}.
\newblock \bibinfo{journal}{\emph{Expert Systems}} \bibinfo{volume}{34},
  \bibinfo{number}{2} (\bibinfo{year}{2017}), \bibinfo{pages}{e12191}.
\newblock


\bibitem[\protect\citeauthoryear{Maas, Hannun, and Ng}{Maas
  et~al\mbox{.}}{2013}]%
        {maas2013rectifier}
\bibfield{author}{\bibinfo{person}{Andrew~L Maas}, \bibinfo{person}{Awni~Y
  Hannun}, {and} \bibinfo{person}{Andrew~Y Ng}.}
  \bibinfo{year}{2013}\natexlab{}.
\newblock \showarticletitle{Rectifier nonlinearities improve neural network
  acoustic models}. In \bibinfo{booktitle}{\emph{Proc. icml}},
  Vol.~\bibinfo{volume}{30}. Citeseer, \bibinfo{pages}{3}.
\newblock


\bibitem[\protect\citeauthoryear{Mikuriya and Cantens}{Mikuriya and
  Cantens}{2020}]%
        {web1}
\bibfield{author}{\bibinfo{person}{Kunio Mikuriya} {and}
  \bibinfo{person}{Thomas Cantens}.} \bibinfo{year}{2020}\natexlab{}.
\newblock \showarticletitle{If algorithms dream of Customs, do customs
  officials dream of algorithms? A manifesto for data mobilization in Customs}.
\newblock \bibinfo{journal}{\emph{World Customs Journal}} \bibinfo{volume}{14},
  \bibinfo{number}{2} (\bibinfo{year}{2020}), \bibinfo{pages}{3--22}.
\newblock


\bibitem[\protect\citeauthoryear{Pozzolo, Boracchi, Caelen, Alippi, and
  Bontempi}{Pozzolo et~al\mbox{.}}{2018}]%
        {Pozzolo}
\bibfield{author}{\bibinfo{person}{Andrea~Dal Pozzolo},
  \bibinfo{person}{Giacomo Boracchi}, \bibinfo{person}{Olivier Caelen},
  \bibinfo{person}{Cesare Alippi}, {and} \bibinfo{person}{Gianluca Bontempi}.}
  \bibinfo{year}{2018}\natexlab{}.
\newblock \showarticletitle{Credit Card Fraud Detection: A Realistic Modeling
  and a Novel Learning Strategy}.
\newblock \bibinfo{journal}{\emph{IEEE Transactions on Neural Networks and
  Learning Systems}} \bibinfo{volume}{29}, \bibinfo{number}{8}
  (\bibinfo{year}{2018}), \bibinfo{pages}{3784--3797}.
\newblock
\showISSN{2162-2388}


\bibitem[\protect\citeauthoryear{Rao, Zhang, Han, Zhang, Min, Chen, Shan, Zhao,
  and Zhang}{Rao et~al\mbox{.}}{2020a}]%
        {rao2020xfraud}
\bibfield{author}{\bibinfo{person}{Susie~Xi Rao}, \bibinfo{person}{Shuai
  Zhang}, \bibinfo{person}{Zhichao Han}, \bibinfo{person}{Zitao Zhang},
  \bibinfo{person}{Wei Min}, \bibinfo{person}{Zhiyao Chen},
  \bibinfo{person}{Yinan Shan}, \bibinfo{person}{Yang Zhao}, {and}
  \bibinfo{person}{Ce Zhang}.} \bibinfo{year}{2020}\natexlab{a}.
\newblock \bibinfo{title}{xFraud: Explainable Fraud Transaction Detection on
  Heterogeneous Graphs}.
\newblock
\newblock
\showeprint[arxiv]{2011.12193}~[cs.LG]


\bibitem[\protect\citeauthoryear{Rao, Zhang, Han, Zhang, Min, Cheng, Shan,
  Zhao, and Zhang}{Rao et~al\mbox{.}}{2020b}]%
        {rao2020suspicious}
\bibfield{author}{\bibinfo{person}{Susie~Xi Rao}, \bibinfo{person}{Shuai
  Zhang}, \bibinfo{person}{Zhichao Han}, \bibinfo{person}{Zitao Zhang},
  \bibinfo{person}{Wei Min}, \bibinfo{person}{Mo Cheng}, \bibinfo{person}{Yinan
  Shan}, \bibinfo{person}{Yang Zhao}, {and} \bibinfo{person}{Ce Zhang}.}
  \bibinfo{year}{2020}\natexlab{b}.
\newblock \bibinfo{title}{Suspicious Massive Registration Detection via Dynamic
  Heterogeneous Graph Neural Networks}.
\newblock
\newblock
\showeprint[arxiv]{2012.10831}~[cs.LG]


\bibitem[\protect\citeauthoryear{Regmi and Timalsina}{Regmi and
  Timalsina}{2018}]%
        {Regmi2018}
\bibfield{author}{\bibinfo{person}{Ram~Hari Regmi} {and}
  \bibinfo{person}{Arun~K. Timalsina}.} \bibinfo{year}{2018}\natexlab{}.
\newblock \showarticletitle{Risk Management in customs using Deep Neural
  Network}. In \bibinfo{booktitle}{\emph{IEEE International Conference on
  Computing, Communication and Security}}. \bibinfo{pages}{133--137}.
\newblock


\bibitem[\protect\citeauthoryear{Rodrigue, Comtois, and Slack}{Rodrigue
  et~al\mbox{.}}{2016}]%
        {rodrigue2016geography}
\bibfield{author}{\bibinfo{person}{Jean-Paul Rodrigue}, \bibinfo{person}{Claude
  Comtois}, {and} \bibinfo{person}{Brian Slack}.}
  \bibinfo{year}{2016}\natexlab{}.
\newblock \bibinfo{booktitle}{\emph{The geography of transport systems}}.
\newblock \bibinfo{publisher}{Routledge}.
\newblock


\bibitem[\protect\citeauthoryear{Schlichtkrull, Kipf, Bloem, Van Den~Berg,
  Titov, and Welling}{Schlichtkrull et~al\mbox{.}}{2018}]%
        {schlichtkrull2018modeling}
\bibfield{author}{\bibinfo{person}{Michael Schlichtkrull},
  \bibinfo{person}{Thomas~N Kipf}, \bibinfo{person}{Peter Bloem},
  \bibinfo{person}{Rianne Van Den~Berg}, \bibinfo{person}{Ivan Titov}, {and}
  \bibinfo{person}{Max Welling}.} \bibinfo{year}{2018}\natexlab{}.
\newblock \showarticletitle{Modeling relational data with graph convolutional
  networks}. In \bibinfo{booktitle}{\emph{European semantic web conference}}.
  Springer, \bibinfo{pages}{593--607}.
\newblock


\bibitem[\protect\citeauthoryear{Shao, Zhao, and Chang}{Shao
  et~al\mbox{.}}{2002}]%
        {shao2002applying}
\bibfield{author}{\bibinfo{person}{Hua Shao}, \bibinfo{person}{Hong Zhao},
  {and} \bibinfo{person}{Gui-Ran Chang}.} \bibinfo{year}{2002}\natexlab{}.
\newblock \showarticletitle{Applying data mining to detect fraud behavior in
  customs declaration}. In \bibinfo{booktitle}{\emph{Proceedings. International
  Conference on Machine Learning and Cybernetics}}, Vol.~\bibinfo{volume}{3}.
  IEEE, \bibinfo{pages}{1241--1244}.
\newblock


\bibitem[\protect\citeauthoryear{Triepels, Daniels, and Feelders}{Triepels
  et~al\mbox{.}}{2018}]%
        {TRIEPELS2018193}
\bibfield{author}{\bibinfo{person}{Ron Triepels}, \bibinfo{person}{Hennie
  Daniels}, {and} \bibinfo{person}{Ad Feelders}.}
  \bibinfo{year}{2018}\natexlab{}.
\newblock \showarticletitle{Data-driven fraud detection in international
  shipping}.
\newblock \bibinfo{journal}{\emph{Expert Systems with Applications}}
  \bibinfo{volume}{99} (\bibinfo{year}{2018}), \bibinfo{pages}{193--202}.
\newblock
\showISSN{0957-4174}


\bibitem[\protect\citeauthoryear{Vanhoeyveld, Martens, and Peeters}{Vanhoeyveld
  et~al\mbox{.}}{2019}]%
        {vanhoeyveld2019customs}
\bibfield{author}{\bibinfo{person}{Jellis Vanhoeyveld}, \bibinfo{person}{David
  Martens}, {and} \bibinfo{person}{Bruno Peeters}.}
  \bibinfo{year}{2019}\natexlab{}.
\newblock \showarticletitle{Customs fraud detection: Assessing the value of
  behavioural and high-cardinality data under the imbalanced learning issue}.
\newblock \bibinfo{journal}{\emph{Pattern Analysis and Applications}}
  (\bibinfo{year}{2019}).
\newblock


\bibitem[\protect\citeauthoryear{Vanhoeyveld, Martens, and Peeters}{Vanhoeyveld
  et~al\mbox{.}}{2020}]%
        {vanhoeyveld2020customs}
\bibfield{author}{\bibinfo{person}{Jellis Vanhoeyveld}, \bibinfo{person}{David
  Martens}, {and} \bibinfo{person}{Bruno Peeters}.}
  \bibinfo{year}{2020}\natexlab{}.
\newblock \showarticletitle{Customs fraud detection}.
\newblock \bibinfo{journal}{\emph{Pattern Analysis and Applications}}
  \bibinfo{volume}{23}, \bibinfo{number}{3} (\bibinfo{year}{2020}),
  \bibinfo{pages}{1457--1477}.
\newblock


\bibitem[\protect\citeauthoryear{Veli{\v{c}}kovi{\'c}, Cucurull, Casanova,
  Romero, Lio, and Bengio}{Veli{\v{c}}kovi{\'c} et~al\mbox{.}}{2017}]%
        {velivckovic2017graph}
\bibfield{author}{\bibinfo{person}{Petar Veli{\v{c}}kovi{\'c}},
  \bibinfo{person}{Guillem Cucurull}, \bibinfo{person}{Arantxa Casanova},
  \bibinfo{person}{Adriana Romero}, \bibinfo{person}{Pietro Lio}, {and}
  \bibinfo{person}{Yoshua Bengio}.} \bibinfo{year}{2017}\natexlab{}.
\newblock \showarticletitle{Graph attention networks}.
\newblock \bibinfo{journal}{\emph{arXiv preprint arXiv:1710.10903}}
  (\bibinfo{year}{2017}).
\newblock


\bibitem[\protect\citeauthoryear{Velickovic, Cucurull, Casanova, Romero,
  Li{\`{o}}, and Bengio}{Velickovic et~al\mbox{.}}{2018}]%
        {gat}
\bibfield{author}{\bibinfo{person}{Petar Velickovic}, \bibinfo{person}{Guillem
  Cucurull}, \bibinfo{person}{Arantxa Casanova}, \bibinfo{person}{Adriana
  Romero}, \bibinfo{person}{Pietro Li{\`{o}}}, {and} \bibinfo{person}{Yoshua
  Bengio}.} \bibinfo{year}{2018}\natexlab{}.
\newblock \showarticletitle{Graph Attention Networks}. In
  \bibinfo{booktitle}{\emph{Proc. of the ICLR}}.
\newblock


\bibitem[\protect\citeauthoryear{Veličković, Fedus, Hamilton, Liò, Bengio,
  and Hjelm}{Veličković et~al\mbox{.}}{2018}]%
        {gnninfomax}
\bibfield{author}{\bibinfo{person}{Petar Veličković},
  \bibinfo{person}{William Fedus}, \bibinfo{person}{William~L. Hamilton},
  \bibinfo{person}{Pietro Liò}, \bibinfo{person}{Yoshua Bengio}, {and}
  \bibinfo{person}{R~Devon Hjelm}.} \bibinfo{year}{2018}\natexlab{}.
\newblock \bibinfo{title}{Deep Graph Infomax}.
\newblock
\newblock
\showeprint[arxiv]{1809.10341}~[stat.ML]


\bibitem[\protect\citeauthoryear{Wang, Lin, Cui, Jia, Wang, Fang, Yu, Zhou,
  Yang, and Qi}{Wang et~al\mbox{.}}{2019a}]%
        {gnn_semifatfraud2019}
\bibfield{author}{\bibinfo{person}{Daixin Wang}, \bibinfo{person}{Jianbin Lin},
  \bibinfo{person}{Peng Cui}, \bibinfo{person}{Quanhui Jia},
  \bibinfo{person}{Zhen Wang}, \bibinfo{person}{Yanming Fang},
  \bibinfo{person}{Quan Yu}, \bibinfo{person}{Jun Zhou},
  \bibinfo{person}{Shuang Yang}, {and} \bibinfo{person}{Yuan Qi}.}
  \bibinfo{year}{2019}\natexlab{a}.
\newblock \showarticletitle{A Semi-Supervised Graph Attentive Network for
  Financial Fraud Detection}. In \bibinfo{booktitle}{\emph{2019 IEEE
  International Conference on Data Mining (ICDM)}}. \bibinfo{pages}{598--607}.
\newblock


\bibitem[\protect\citeauthoryear{Wang, Wen, Wu, Huang, and Xion}{Wang
  et~al\mbox{.}}{2019b}]%
        {fdgars2019}
\bibfield{author}{\bibinfo{person}{Jianyu Wang}, \bibinfo{person}{Rui Wen},
  \bibinfo{person}{Chunming Wu}, \bibinfo{person}{Yu Huang}, {and}
  \bibinfo{person}{Jian Xion}.} \bibinfo{year}{2019}\natexlab{b}.
\newblock \showarticletitle{FdGars: Fraudster Detection via Graph Convolutional
  Networks in Online App Review System}. In \bibinfo{booktitle}{\emph{Companion
  Proceedings of The 2019 World Wide Web Conference}}.
  \bibinfo{pages}{310–316}.
\newblock


\bibitem[\protect\citeauthoryear{Wang, He, Feng, Nie, and Chua}{Wang
  et~al\mbox{.}}{2018}]%
        {wang2018www}
\bibfield{author}{\bibinfo{person}{Xiang Wang}, \bibinfo{person}{Xiangnan He},
  \bibinfo{person}{Fuli Feng}, \bibinfo{person}{Liqiang Nie}, {and}
  \bibinfo{person}{Tat-Seng Chua}.} \bibinfo{year}{2018}\natexlab{}.
\newblock \showarticletitle{{TEM: T}ree-enhanced embedding model for
  explainable recommendation}. In \bibinfo{booktitle}{\emph{WWW}}.
  \bibinfo{pages}{1543--1552}.
\newblock


\bibitem[\protect\citeauthoryear{West and Bhattacharya}{West and
  Bhattacharya}{2016}]%
        {WEST201647}
\bibfield{author}{\bibinfo{person}{Jarrod West} {and} \bibinfo{person}{Maumita
  Bhattacharya}.} \bibinfo{year}{2016}\natexlab{}.
\newblock \showarticletitle{Intelligent financial fraud detection: A
  comprehensive review}.
\newblock \bibinfo{journal}{\emph{Computers \& Security}}  \bibinfo{volume}{57}
  (\bibinfo{year}{2016}), \bibinfo{pages}{47--66}.
\newblock
\showISSN{0167-4048}


\bibitem[\protect\citeauthoryear{Wright}{Wright}{2019}]%
        {Ranger}
\bibfield{author}{\bibinfo{person}{Less Wright}.}
  \bibinfo{year}{2019}\natexlab{}.
\newblock \bibinfo{title}{Ranger - a synergistic optimizer.}
\newblock
  \bibinfo{howpublished}{\url{https://github.com/lessw2020/Ranger-Deep-Learning-Optimizer}}.
\newblock


\bibitem[\protect\citeauthoryear{Ying, He, Chen, Eksombatchai, Hamilton, and
  Leskovec}{Ying et~al\mbox{.}}{2018}]%
        {pinsage}
\bibfield{author}{\bibinfo{person}{Rex Ying}, \bibinfo{person}{Ruining He},
  \bibinfo{person}{Kaifeng Chen}, \bibinfo{person}{Pong Eksombatchai},
  \bibinfo{person}{William~L. Hamilton}, {and} \bibinfo{person}{Jure
  Leskovec}.} \bibinfo{year}{2018}\natexlab{}.
\newblock \showarticletitle{Graph Convolutional Neural Networks for Web-Scale
  Recommender Systems}. In \bibinfo{booktitle}{\emph{Proceedings of the 24th
  ACM SIGKDD International Conference on Knowledge Discovery \&and Data
  Mining}}. \bibinfo{pages}{974–983}.
\newblock


\bibitem[\protect\citeauthoryear{Yoon, Zhang, Jordon, and van~der Schaar}{Yoon
  et~al\mbox{.}}{2020}]%
        {yoon2020vime}
\bibfield{author}{\bibinfo{person}{Jinsung Yoon}, \bibinfo{person}{Yao Zhang},
  \bibinfo{person}{James Jordon}, {and} \bibinfo{person}{Mihaela van~der
  Schaar}.} \bibinfo{year}{2020}\natexlab{}.
\newblock \showarticletitle{Vime: Extending the success of self-and
  semi-supervised learning to tabular domain}.
\newblock \bibinfo{journal}{\emph{Advances in Neural Information Processing
  Systems}}  \bibinfo{volume}{33} (\bibinfo{year}{2020}),
  \bibinfo{pages}{11033--11043}.
\newblock


\bibitem[\protect\citeauthoryear{Zhu, Luo, Li, Bu, Zhou, Zhang, and Lu}{Zhu
  et~al\mbox{.}}{2020}]%
        {gnnminigraph}
\bibfield{author}{\bibinfo{person}{Yong-Nan Zhu}, \bibinfo{person}{Xiaotian
  Luo}, \bibinfo{person}{Yu-Feng Li}, \bibinfo{person}{Bin Bu},
  \bibinfo{person}{Kaibo Zhou}, \bibinfo{person}{Wenbin Zhang}, {and}
  \bibinfo{person}{Mingfan Lu}.} \bibinfo{year}{2020}\natexlab{}.
\newblock \showarticletitle{Heterogeneous Mini-Graph Neural Network and Its
  Application to Fraud Invitation Detection}. In \bibinfo{booktitle}{\emph{2020
  IEEE International Conference on Data Mining (ICDM)}}.
  \bibinfo{pages}{891--899}.
\newblock


\end{thebibliography}

\appendix
\section{Appendix}

\subsection{Algorithm and Complexity Analysis}
\begin{algorithm}[ht!]
\SetKwInput{KwInput}{Input}                
\SetKwInput{KwOutput}{Output}              
\SetAlgoLined
\KwInput{Graph $\mathcal{G(V,E)}$; node feature matrix $\mathbf{D}$; depth $K$; learnable weight matrices $\Theta = \{ \mathbf{W}_1^{(k)}, \mathbf{W}_2^{(k)}, \forall k \in \{ 1,...,K\}$

, $\mathbf{r}_1, \mathbf{r}_2 \}$;
labeled data $L = (x_i, y_i^{cls}, y_i^{rev})$} 
 \KwOutput{Trained weights $\Theta$}

 Initialize $\mathbf{D}$ with zero matrix
 Train XGBoost model with $L$ using $(x_i, y_i^{cls})$ \\ 
 Update transaction feature in $\textbf{D}$ with leaf indices from XGBoost \\
 $s_v^{(0)} \gets d_v,  \forall d_v \in \mathbf{D}$ \\ 
\For{$ \text{step} \in\{1,2,...,\text{pretraining epochs}$\}}{
    \For{$k=1..K$}
    {
        \For{$v \in \mathcal{V}$}
        {
            Update $s_v^{(k)}$ with E.q.~\ref{eq:aggregation}
        }
    }
    Update $W_1^{(k)}, W_2^{(k)}, \forall k \in \{ 1,...,K\}$ with E.q.~\ref{loss:pretrain}
}
obtain the set of labeled nodes $\mathcal{V}_L \in\mathcal{V}$\\
\For{$ \text{step} \in\{1,2,...,\text{finetuning epochs}$\}}{
    \For{$k=1..K$}
    {
        \For{$v \in \mathcal{V}_L$}
        {
            Update $s_v^{(k)}$ with E.q.~\ref{eq:aggregation}
        }
    }
    Obtain predictions $\hat{y}^{cls}_{i}$ and $\hat{y}^{rev}_{i}$ with E.q.~\ref{eq:prediction} \\
    Update $\Theta$ with E.q.~\ref{eq:loss}
}
         
output trained weight $\Theta$
 \caption{Training algorithm for \model{}}
 \label{algor}
\end{algorithm}
For a better understanding of the training procedure of \model{}, we elaborate the pseudo-code in Algorithm~\ref{algor}. First, we train the XGBoost model and obtain the leaf indices in lines 1 and 2. Afterwards, line 4 to 10 describes the pretraining stage and learn the weight matrices $\{ \mathbf{W}_1^{(k)}, \mathbf{W}_2^{(k)}$. Finally, lines 13 to 22 presents the fine-tuning stage that trains all the parameters $\Theta$ with Eq.~\ref{eq:loss} and outputs the trained parameters.

For space complexity, we have $O(|\mathcal{V}| + |\mathcal{E}|)$ as we need to keep all the nodes and edges in track. Since each transaction is connected to $l$ virtual nodes in maximum, the size of $\mathcal{E}$ could be replaced with $n\times l$, where $n$ is the number of transactions. Thus the space complexity could be further derive as $O(n(l+2))$, where $|\mathcal{V}|=n(l+1), |\mathcal{E}|=nl$ 
For time complexity, the layer-wise message passing rule is the main operation. Without the sampling method mentioned in Sec.~\ref{method:local_subgraph}, the memory and expected runtime of a single batch is unpredictable and, in the worst case, could be $O(|\mathcal{V}|)$. In contrast, as we sample a fixed number of $T_i$ in $k$-th propagation layer, the per-batch space and time complexity for \model{} is fixed at $O(\prod_{i=1}^K T_i)$.

\subsection{Hyperparameters}
\model{} has two sets of hyperparameters at the macro level: GBDT and GNN.
For GBDT, we use $100$ trees and a maximum depth of $4$ (same as DATE). For unsupervised and semi-supervised GNNs, the number of GNN layers are kept at $2$ with neighborhood sample sizes $T_1=25, T_2=10$ .
The embedding dimension is searched in $\{8, 16, 32, 64\}$ and eventually set to 32, and $\lambda$ is $1e-4$. The balancing factor $\alpha$ is searched in $\{ 0.1,1,10\}$ and finally set as 10.
Similarly, the batch size for the GNN models is $512$. Additionally, the learning rate is searched in $\{0.05,0.01,0.005,0.001\}$ and set at $0.005$ for C-data and A-data, and 0.05 for B data. 
For XGBoost model, we used the default 100 trees and max depth 4 and kept it same for all other experiments. We pre-trained Tabnet in an unsupervised fashion with an entire training set - including both labeled and unlabeled data and then fine-tuned the model with labeled data. Embedding dimensions for Tabnet and DATE were searched in  $\{8, 16, 32, 64\}$, and learning rate in  $\{0.0001, 0.001, 0.01\}$. For VIME, we tune the learning rate in $\{0.0001, 0.001, 0.01\}$ with Adam optimizer and the hidden dimension in $\{256,512,1024\}$ and select the best-performing parameter on validation set. 

\subsection{Design Decisions}

\begin{table}[!ht]
\caption{\model{} Performance under different design choices}
\label{tab:design-results}
\begin{tabular}{|c|c|c|c|c|c|c|}

\hline
\multicolumn{7}{|c|}{\textbf{A-Data}} \\ \hline
& \multicolumn{3}{c|}{\textbf{n = 1\%}} & \multicolumn{3}{c|}{\textbf{n=5\%}} \\ \hline
Model & Pre. & Rec. & Rev. & Pre. & Rec. & Rev.   \\ \hline
GNN\textsubscript{dense}  & 0.030 & 0.024 & 0.037& 0.017 & 0.069 & 0.100 \\ \hline
GNN\textsubscript{deep}  & 0.033 & 0.026 & \textbf{0.045} & 0.022 & 0.085 & 0.128 \\ \hline
\model{} & \textbf{0.038} & \textbf{0.030} & 0.044 & \textbf{0.025} & \textbf{0.097} & \textbf{0.182}  \\ \hline

\multicolumn{7}{|c|}{\textbf{B-Data}} \\ \hline
GNN\textsubscript{dense}  & 0.211 & 0.085 & 0.123  & \textbf{0.148} & \textbf{0.299} & \textbf{0.295}\\ \hline
GNN\textsubscript{deep}  & 0.195 & 0.079 & 0.120& 0.079 & 0.158 & 0.227 \\ \hline
\model{} & \textbf{0.245} & \textbf{0.099} & \textbf{0.131}& 0.118 & 0.238 & 0.283 \\ \hline

\multicolumn{7}{|c|}{\textbf{C-Data}} \\ \hline
GNN\textsubscript{dense}  & 0.837 & 0.088 & \textbf{0.174}& 0.556 & 0.293 & 0.457 \\ \hline
GNN\textsubscript{deep}  & \textbf{0.876} & 0.092 & 0.165 & \textbf{0.583} & \textbf{0.308} & \textbf{0.479} \\ \hline
\model{} & 0.869 & \textbf{0.092} & 0.170& 0.535 & 0.282 & 0.445 \\ \hline

\end{tabular}
\end{table}
In this section, we explore the effect of choice of most noteworthy hyper-, and other parameters on the model performance. For instance, we utilize two of the categorical features in the data to build the graph structure from the customs data. It would be interesting to evaluate the effect of the inclusion of other categorical features as well. Other parameters such as layer depth of GNNs are directly linked to the overall architecture and hence, could directly affect the model performance. We vary these parameters as follows:

\begin{itemize}[leftmargin=*]
    \item \textbf{GNN depth}: Even though we designed the graph structure in our model to be \textit{less dense}, the number of nodes and edges can nonetheless run into the order of multi-millions. To reduce the training time, and to avoid the over-smoothing of GNNs, we restrict the layer depth of GNNs to be 2.  In this experiment, we explore whether increasing the layer depth would offer any additional performance gains.
    \item \textbf{GNN{\textsubscript{dense}}}: During the GBDT runs, we noticed that some of the designed features designed during the feature engineering phase prove to be most effective in building the GBDT model. In this experimental setup, we utilize additional categorical information (a combination of importer and office IDs) to build a \textit{denser} graph. 
\end{itemize}

\subsection{Supervised Learning}

\begin{table}[!ht]
\caption{Model performance with a 100\% inspection rate (supervised setting).}
\label{tab:sup-results}
\begin{tabular}{|c|c|c|c|c|c|c|}
\hline
\multicolumn{7}{|c|}{\textbf{A-Data}} \\ \hline
& \multicolumn{3}{c|}{\textbf{n = 1\%}} & \multicolumn{3}{c|}{\textbf{n=5\%}} \\ \hline
Model & Pre. & Rec. & Rev. & Pre. & Rec. & Rev.   \\ \hline
XGB  & 0.077 & 0.061 & 0.015& 0.051 & 0.201 & 0.341 \\ \hline
Tabnet  & 0.054 & 0.036 & 0.070 & 0.053 & 0.196 & 0.339\\ \hline
DATE  & 0.081 & 0.064 & \textbf{0.129} & 0.056 & 0.219 & \textbf{0.378} \\ \hline
\model{} & \textbf{0.087} & \textbf{0.069} & 0.121 & \textbf{0.056} & \textbf{0.222} & 0.334 \\ \hline

\multicolumn{7}{|c|}{\textbf{B-Data}} \\ \hline
XGB  & 0.896 & 0.362 & 0.249 & 0.373 & 0.754 & 0.604 \\ \hline
Tabnet  & 0.921 & 0.372 & 0.245 & 0.387 & 0.782 & \textbf{0.675} \\ \hline
DATE  & 0.937 & 0.379 & 0.265 & \textbf{0.410} & \textbf{0.829} & 0.619 \\ \hline
\model{} & \textbf{0.963} & \textbf{0.389} & \textbf{0.271} & 0.396 & 0.801 & 0.670 \\ \hline

\multicolumn{7}{|c|}{\textbf{C-Data}} \\ \hline
XGB  & 0.844 & 0.089 & 0.158 & 0.539 & 0.284 & 0.421 \\ \hline
Tabnet  & 0.824 & 0.070 & 0.157 & 0.500 & 0.261 & 0.402  \\ \hline
DATE  & 0.810 & 0.085 & 0.178  & 0.532 & 0.280 & 0.426 \\ \hline
\model{} & \textbf{0.912} & \textbf{0.096} & \textbf{0.188} & \textbf{0.587} & \textbf{0.310} & \textbf{0.450} \\ \hline

\end{tabular}
\end{table}
This setting pertains to the availability of 100\% of the ground-truth labels for the historical customs data. Availability of ground-truth labels for the entirety of the customs transactions is relatively rare and infeasible. Our main focus in the paper is semi-supervised setting but nonetheless, we test the model performance in full supervised setting as we have fully labeled import transactions from three countries. Table~\ref{tab:sup-results} presents the performance comparison with baselines. \model{} outperforms all the baselines consistently, inspite of the fact that baselines are already performing quite well. As an example, for B-data, the Pre@1 increases from DATE's already impressive 0.937 to 0.963 and Rec@1 from 0.379 to 0.389. Furthering performance gain from such a level is quite impressive. For C-data, we observe a gain of about 12\% and 13\% for Pre.@1 and Rec.@1 respectively. For A-data, the overall behavior remains same as that of semi-supervised, with results being same or slightly better than a strong baseline.

\subsection{Classification and regression losses}
We define the losses for classification ($\mathcal{L}_{cls}$) and regression loss $\mathcal{L}_{rev}$ for dual task leanring as follows:

\begin{equation}
\label{eq:l_cls}
\begin{split}
    \mathcal{L}_{cls} &= - \frac{1}{n} \sum_i  y^{cls}_{i} \log(\hat{y}^{cls}_{i}(s_m^{(k)}) + (1-y^{cls}_{i}) \log(1-\hat{y}^{cls}_{i}(s_m^{(k)}), \\
    \mathcal{L}_{rev} &= \frac{1}{n} \sum_i \left(y^{rev}_{i} - \hat{y}^{rev}_{i}(s_m^{(k)})\right)^2,
\end{split}
\end{equation}
where $y^{cls}_{i}$ and $y^{rev}_{i}$ are the ground-truth illicitness class and raised revenue of transactions $t_i$, respectively, $\lambda$ is the regularization hyperparameter to prevent overfitting, and $n$ is the number of training samples. 
Further, the revenue collection can be defined as: 
\begin{equation}
\label{eq:rev_c}
\begin{aligned}
    \widehat{rev}^{} &=  \frac{\sum_i{S_{i}y_{i}^{cls}\zeta_{\hat{y}_{i}^{{cls}_{}}}}}{\sum_i{S_{i}y_{i}^{cls}}}
\end{aligned}
\end{equation}
where $\widehat{rev}$ is the total revenue collected by inspecting a certain fraction of items over a set of test items $i$, $S_i$ is the penalty surcharge an importer has to pay on \textit{that} transaction if a fraudulent entry is detected, $y_{i}^{cls}$ is the ground-truth label for a transaction, and $\zeta_{\hat{y}_{i}^{{cls}_{}}}$ is a boolean flag valued true if the model's prediction $\hat{y}_{i}^{cls}$ equates to a true positive, and false otherwise.

\end{document}